\ifdefined\pdfobjcompresslevel\pdfobjcompresslevel=0\fi
\documentclass{fairmeta}
\microtypesetup{expansion=false}
\usepackage{graphicx}
\usepackage{amsmath,amssymb}
\usepackage{booktabs}
\usepackage{multirow}
\usepackage{xspace}
\usepackage{microtype}
\usepackage{tabularx}
\usepackage{tcolorbox}
\tcbuselibrary{breakable}
\newcounter{vlnalgorithm}
\definecolor{promptblue}{HTML}{0070C0}
\usepackage{array}
\usepackage{needspace}
\usepackage{colortbl}
\definecolor{navilahighlight}{rgb}{0.92,0.97,0.85}
\newcommand{\method}{\textbf{ZeroVLN}\xspace}
\newcommand{\tool}[1]{\texttt{#1}}
\definecolor{cvprblue}{rgb}{0.21,0.49,0.74}
\hypersetup{breaklinks,colorlinks,allcolors=metablue,pdftitle={ZeroVLN: Learning to Navigate without Supervised Fine-Tuning of the VLM Brain}}
\title{ZeroVLN: Learning to Navigate without Supervised Fine-Tuning of the VLM Brain}
\author[1,2,\dagger]{Zixuan Huang}
\author[2,3,\dagger]{Yang Zhou}
\author[2]{Guli Zhang}
\author[2]{Sunzhu Li}
\author[1]{Hongyan Xie}
\author[1]{Zehao Chen}
\author[1]{Fuzhen Zhuang}
\author[2]{Kaixuan Wang}
\author[2]{Xiaozhi Chen}
\author[1,*]{Yikun Ban}
\author[1,*]{Deqing Wang}
\renewcommand{\affiliationlist}{%
\affiliationformat[1]{Beihang University},
\affiliationformat[2]{Zhuoyu Technology, Shenzhen, China},
\affiliationformat[3]{Zhejiang University}%
}
\contribution[\dagger]{Equal contribution}
\contribution[*]{Corresponding author}
\date{2026.11.17}
\correspondence{\email{huang\_zx@buaa.edu.cn}}
\code{https://github.com/hzx122/ZeroVLN}
\models{https://huggingface.co/hzxllll/ZeroVLN}
\abstract{

Representative vision-language navigation (VLN) methods typically couple a vision-language model (VLM) brain with a learnable action head and fine-tune the VLM brain on navigation-domain data to predict geometric goals.
However, such adaptation can substantially alter base VLM weights and degrade general visual capabilities.
We introduce \method, a slow--fast navigation framework that achieves effective navigation with zero navigation-domain SFT of the VLM brain.
Rather than teaching the brain to predict geometric goals, we equip it with a semantic tool-calling interface for conveying movement decisions to an action expert in natural language. 
We first adapt the action expert to carry out these movement requests, then jointly improve the brain's movement decisions and the action expert's continuous execution through closed-loop reinforcement learning (RL).
Trained on the same navigation episodes, \method{} substantially outperforms representative SFT-based hierarchical VLN methods on standard VLN-CE (continuous environments) benchmarks, while inducing only minor changes to the brain's weights and largely preserving its general capabilities. 
To our knowledge, \textbf{\method{} is the first slow--fast VLN framework with zero navigation-domain SFT of the VLM brain, and the first to jointly optimize its slow and fast policies through closed-loop agentic RL in VLN-CE.}

}

\begin{document}
\maketitle

\addtocontents{toc}{\protect\setcounter{tocdepth}{-1}}
\section{Introduction}

\begin{figure}[!htb]
\centering
\includegraphics[width=0.5\textwidth]{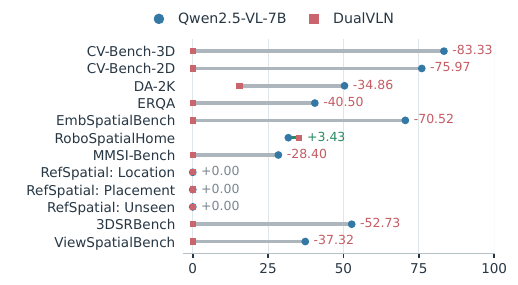}
\caption{\textbf{General Capability after Navigation Adaptation.} Compared with the base Qwen2.5-VL-7B model, DualVLN loses substantial accuracy on most visual and spatial benchmarks, motivating navigation learning without brain SFT.}
\label{fig:dualvln_motivation}
\end{figure}

VLN requires an embodied agent to connect movement decisions with precise movement~\cite{cheng2024navila,qi2025vln,anderson2018r2r,ku2020rxr,krantz2020beyond,wang2025vlnpe}. Representative methods adapt a general-purpose VLM through navigation-domain supervision to predict geometric goals for an action expert~\cite{wei2025dualvln,gong2026abotn1} or generate numerical movement commands in a predefined format~\cite{cheng2024navila}.
However, we observe substantial displacement of the base VLM's weights and degraded performance of general visual capability following this adaptation, as shown in Figure~\ref{fig:dualvln_motivation}.

Inspired by DeepSeek-R1-Zero~\cite{deepseekai2025r1}, which demonstrates that reasoning capabilities can emerge through RL directly from a base language model without prior SFT, 
we introduce \method{}, the first slow--fast navigation framework that learns navigation through environment interaction with \textbf{zero navigation-domain SFT of the VLM brain}. 
Rather than adapting the brain to predict geometric goals, we introduce a semantic tool interface that connects it to a trainable flow-matching action expert~\cite{yi2026flowpolicygradients}. 
Through natural-language tool calls, \textbf{the brain conveys movement decisions to the action expert, which translates them into continuous motion.}

Within this framework, we first \textbf{construct reference trajectories from the brain's own movement requests}, collected through online rollouts with the brain frozen. To ground these requests in executable motion, we design a privileged trajectory planner that leverages a semantic map derived from the scene mesh~\cite{miao2025sage}. 
The resulting reference trajectories are paired with their originating requests to provide semantically aligned supervision for the action expert.

We then \textbf{jointly optimize the brain's movement decisions and the action expert's continuous execution} through multi-round agentic RL~\cite{feng2026group,zheng2026pgpo}.
To address sparse success signals in challenging long-horizon navigation tasks, we \textbf{design complementary process and outcome rewards} that distinguish useful progress even when task completion fails.

Beyond conventional slow--fast frameworks, our design provides two key benefits: 
\textbf{(1)} the semantic interface leverages the brain's native  capabilities for flexible tool use and feedback-driven recovery~\cite{yu2025correctnav,li2026agenticnav}; 
\textbf{(2)} Rather than specializing the model for navigation through SFT, \textbf{\method builds upon its existing language and navigation decision-making capabilities}, using RL to enhance navigation performance while largely preserving its general-purpose abilities.

Our main contributions are summarized as follows:
\begin{itemize}
\item We introduce \method{}, the first slow--fast VLN framework that connects a VLM brain with an action expert policy through a semantic tool interface, achieving \textbf{zero navigation-domain SFT of the VLM brain}.
\item We develop a two-stage training paradigm that first aligns continuous execution with the brain's semantic requests through \textbf{request-grounded supervision and execution-prefix learning}, then jointly optimizes the slow and fast policies through \textbf{closed-loop agentic RL}.
\item Across mainstream VLN-CE benchmarks, \method{} substantially outperforms representative SFT-based baselines trained on the same navigation episodes, with only \textbf{0.01497\% VLM weight displacement} while largely preserving general visual capabilities.
\end{itemize}

\section{Architecture}
\label{sec:architecture}

\begin{figure}[!htb]
\centering
\includegraphics[width=0.9\textwidth]{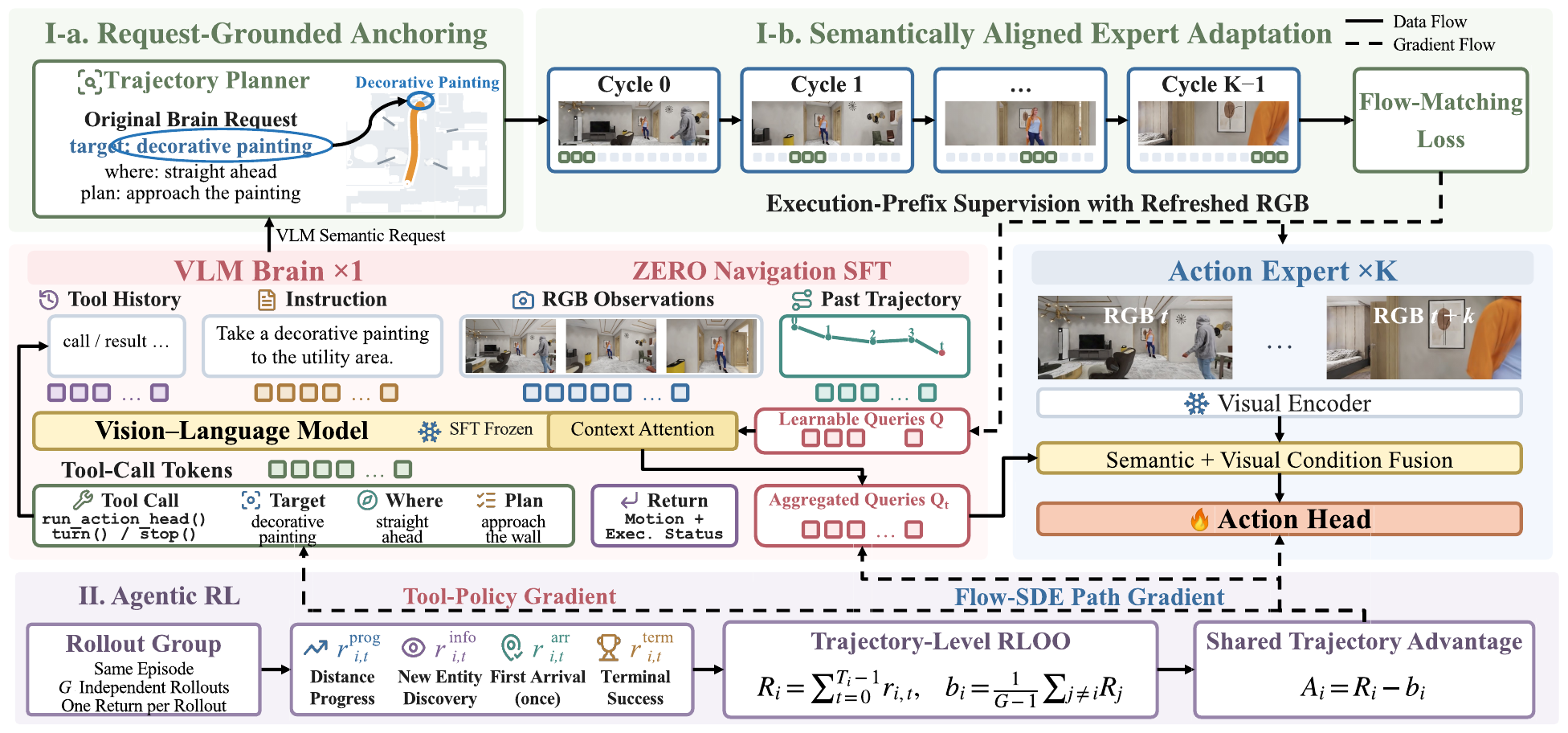}
\caption{\textbf{Framework of ZeroVLN.} The brain delegates semantic requests to an action expert. Request-grounded supervision first adapts the expert; joint agentic RL then improves both policies without navigation-domain SFT of the brain.}

\label{fig:architecture}
\end{figure}

\paragraph{Semantic Tool Calls and Contextual Queries.}

As illustrated in Figure~\ref{fig:architecture}, we separate movement decisions from continuous motion execution through a semantic tool-calling interface. Rather than predicting geometric goals, the VLM brain makes movement decisions via \tool{run\_action\_head(target, where, plan)}, specifying a target, its relative location, and a local plan for the action expert. The brain can also inspect the scene through \tool{turn(direction, degrees)} or terminate navigation through \tool{stop(reason)}, with prompts and tool schemas detailed in Appendix~\ref{sec:navigation_system_prompt}.

At slow decision round $t$, the brain policy $\pi_\theta$ generates a tool call $y_t$ from multimodal context $x_t$, including the navigation instruction, RGB observations, tool history, and trajectory trace. For movement calls, $q_t$ denotes the natural-language arguments of $y_t$. To convey the request and its navigation context to the action expert, we concatenate learnable query embeddings $Q$ with the embedded context and tool call. The VLM contextualizes these queries, whose final-layer hidden states $Q_t$ provide the action expert with semantic conditioning~\cite{wei2025dualvln}:

{\small
\begin{equation}
Q_t=\left[
\operatorname{VLM}_{\theta}
\left(\operatorname{Embed}(x_t,y_t)\oplus Q\right)
\right]_Q.
\label{eq:context_queries}
\end{equation}
}

\begin{figure}[!htb]
\centering
\includegraphics[width=0.8\textwidth]{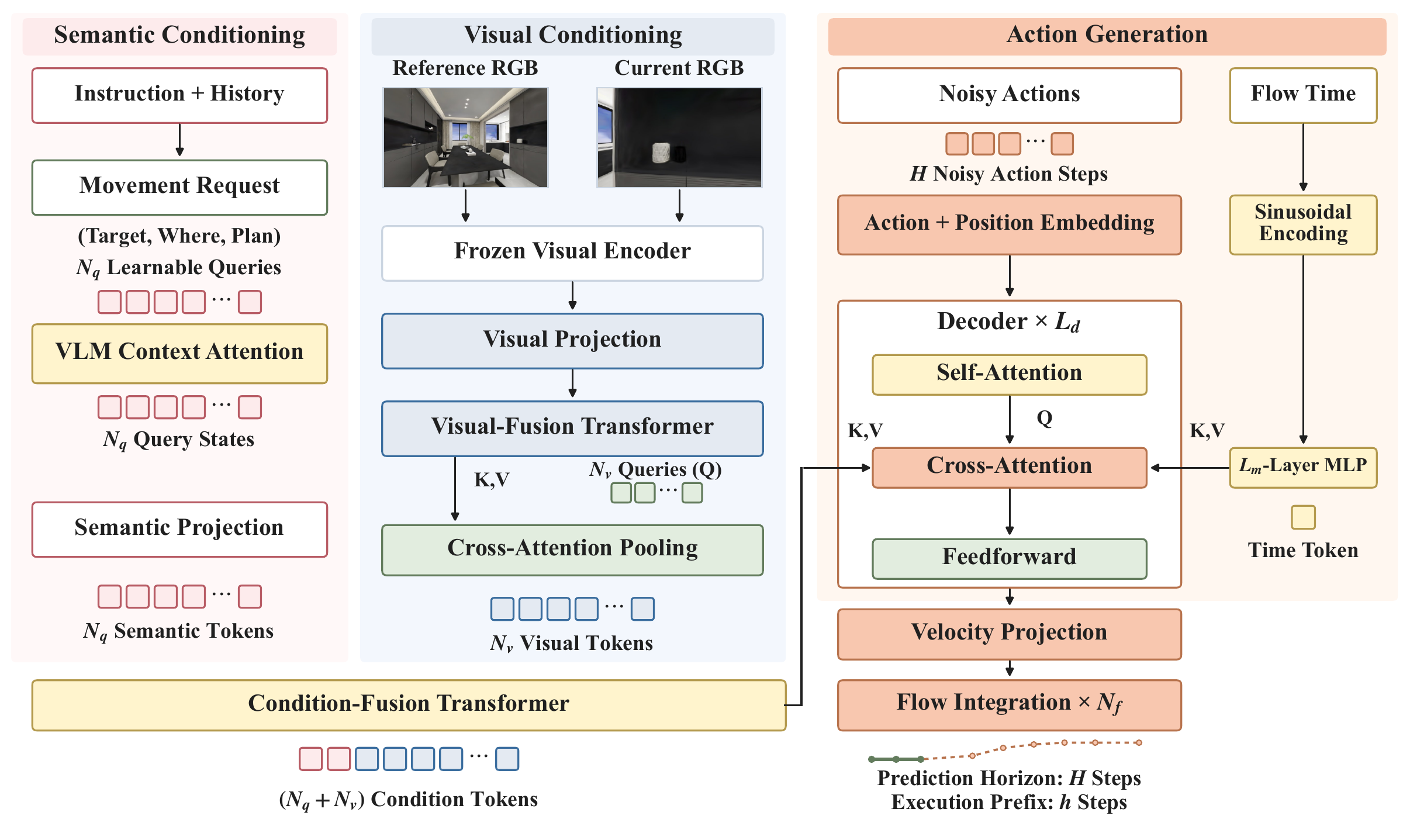}
\caption{\textbf{Action Expert Architecture.} Semantic queries ($N_q$) and visual tokens ($N_v$) condition action generation. $N_f$ flow steps predict an $H$-step sequence; only its $h$-step prefix is executed before visual replanning.}
\label{fig:action_expert_architecture}
\end{figure}

\paragraph{Visual Execution at a Faster Timescale.}

While the brain specifies the intended movement, the action expert executes the call over multiple fast cycles. Throughout each call, the semantic condition $Q_t$ and reference image $I_t^{\mathrm{ref}}$ remain fixed, while RGB observations $I_{t,k}$ are refreshed at each cycle~\cite{wei2025dualvln}. A frozen visual encoder and trainable fusion module combine both images into visual features $F_{t,k}$. The semantic adapter $P_1$ projects $Q_t$ into the visual conditioning space and adds a semantic-channel embedding, while $P_2$ adds a visual-channel embedding to $F_{t,k}$. Concatenating the two streams forms the action expert's conditioning input:

{\small
\begin{equation}
C_{t,k}=P_1(Q_t)\oplus P_2(F_{t,k}).
\label{eq:expert_condition}
\end{equation}
}

Conditioned on $C_{t,k}$, the continuous-action expert $\pi_\phi$ integrates the flow velocities predicted by $f_\phi$ to generate an $H$-step action sequence. After inverse normalization, each action specifies relative planar displacements and a heading change in the local frame defined by the agent's pose at the start of each cycle. At each cycle, we execute only the first $h\leq H$ steps, then refresh the observation and predict the next sequence, adapting motion while preserving the brain's intent. After each movement call, visual and trajectory feedback, including blocked motion or collisions, allows the brain to continue, inspect the scene, revise its request, or stop. Figure~\ref{fig:action_expert_architecture} details this execution path: $N_q$ semantic queries and $N_v$ visual tokens condition an $L_d$-layer decoder, with an $L_m$-layer flow-time MLP and $N_f$ flow integration steps. Appendix~\ref{sec:visual_conditioning} provides further details on the action expert's architecture and execution.

\section{Learning to Execute Semantic Requests}
\label{sec:expert_adaptation}

\paragraph{Request-Grounded Anchoring.}

Our semantic interface also changes how action expert supervision must be constructed.
Geometric-interface VLN frameworks can derive training conditions directly from scene meshes or ground-truth trajectories~\cite{wei2025dualvln,gong2026abotn1,an2023etpnav}, whereas our action expert follows natural-language requests issued online by the brain itself.
Directly constructing semantic requests may not match the brain's actual request distribution, leaving the action expert exposed to out-of-distribution instructions at inference.
Generating requests online along fixed trajectories also leaves the action labels independent of the requested movement.
We report results for these alternatives, including CAST-inspired augmentation~\cite{glossop2025cast} and request transfer in Appendix~\ref{sec:anchoring_alternatives}.

\begin{figure}[!htb]
\centering
\includegraphics[width=0.85\textwidth]{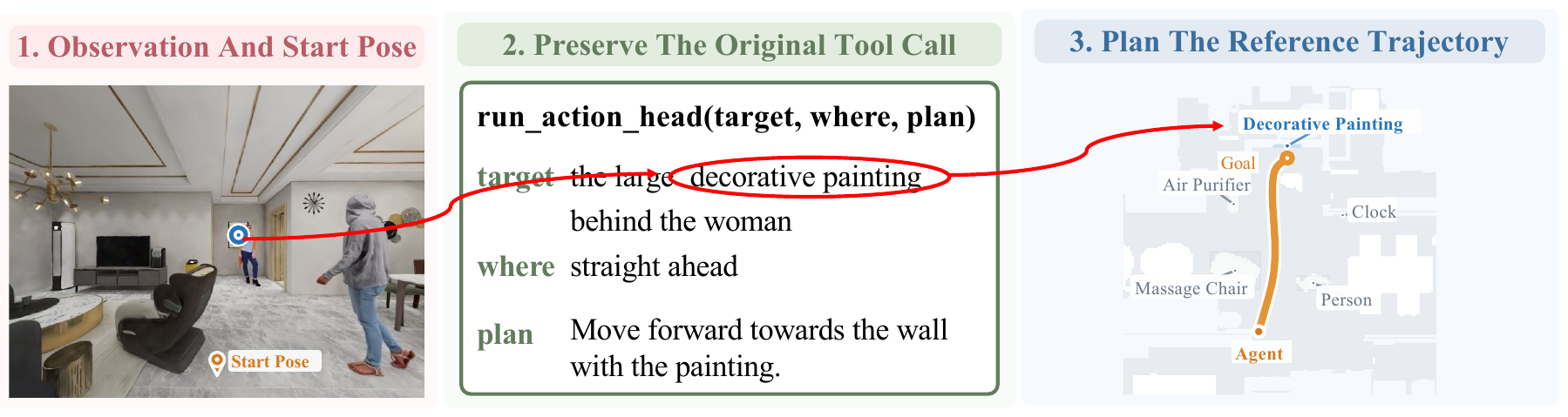}
\caption{\textbf{Request Grounding And Reference Trajectory Generation.} The trajectory planner maps the original movement request to a scene instance and reachable endpoint. Red curves mark grounding correspondences. Semantic maps and mesh geometry are used only during training.}
\label{fig:open_loop_training}
\end{figure}

Built upon these insights, we propose a request-grounded data construction strategy that \textbf{generates trajectories from the brain's actual requests}.
To realize this, we design a privileged semantic planner that uses a mesh-derived semantic map~\cite{miao2025sage} and training-only geometry to ground each request \(q_t\) at the agent's current state \(s_t\) into a reachable local goal \(g_t^\star\), from which a reference trajectory \(\rho_t\) is planned.
The resulting execution-verified trajectories are paired with their originating requests to provide semantically aligned supervision for action expert adaptation.
Further details of the semantic planner are provided in Appendix~\ref{sec:semantic_teacher_details}.

\paragraph{Execution-Prefix Supervision.}

These request-grounded trajectories provide aligned training targets, but supervision must also match the action expert's execution criteria. 
For each verified trajectory $\rho_t$, we construct $H$-step reference action chunks $U_{t,k}^\star\in\mathbb R^{H\times d_a}$ across $K$ fast execution cycles, with $d_a$ action components per step, and pair them with the corresponding conditions $C_{t,k}$. 
Additional details on action construction are provided in Appendix~\ref{sec:action_target_details}.

Existing action-chunking methods  supervise the full prediction horizon despite executing only a short prefix before replanning~\cite{wei2025dualvln,chi2025diffusion,black2024pi_0}. Once new observations arrive, the remaining predictions are discarded, creating a mismatch between training and execution. We therefore \textbf{retain full-horizon prediction while restricting flow-matching supervision to the executed $h$-step prefix}.
At flow time $u$ sampled from a logit-normal distribution, we interpolate the full reference chunk with Gaussian noise $\epsilon\sim\mathcal N(0,I)$, but compute the loss only over its first $h$ actions:

{\small
\begin{equation}
U_{t,k}^{(u)}
=(1-u)U_{t,k}^\star+u\epsilon,\qquad
\mathcal L_{\mathrm{FM}}
=\mathbb E\Bigg[
\frac{1}{K}\sum_{k=1}^{K}
\frac{1}{hd_a}\sum_{j=1}^{h}\left\|
\left[
f_\phi(U_{t,k}^{(u)},u,C_{t,k})
-(\epsilon-U_{t,k}^\star)
\right]_j
\right\|_2^2
\Bigg].
\label{eq:prefix_fm}
\end{equation}
}

During this stage, the VLM brain and visual encoder remain frozen, while the learnable queries $Q$ and action expert parameters $\phi$ are optimized.

\section{Joint Slow--Fast Optimization via Agentic RL}
\label{sec:credit}

\subsection{Trajectory-Level Credit Assignment}
\label{sec:credit_framework}

Supervised adaptation enables the action expert policy $\pi_\phi$ to execute individual movement requests, 
while successful navigation also depends on the VLM brain's ability to make movement decisions through semantic tool calls.

To jointly improve movement decisions and continuous execution, we optimize the VLM brain policy $\pi_\theta$ and action expert policy $\pi_\phi$ using navigation feedback through closed-loop agentic RL~\cite{feng2026group,zheng2026pgpo}.
Accurately estimating round-level advantages is challenging because task success depends on decisions and execution across multi-round interaction with the environment. 
We instead adopt trajectory-level credit assignment to evaluate their combined contribution to navigation success.

For each navigation task, we sample $G$ independent long-horizon rollouts, with rollout $i$ comprising $T_i$ rounds of tool calls, corresponding action executions, and per-round environment feedback, which is used to compute the reward for each round. Using RLOO advantage estimation~\cite{ahmadian2024back,shao2024deepseekmath}, we compute the advantage $A_i$ by comparing its cumulative return $R_i$ with the mean return of the remaining $G-1$ rollouts:

{\small
\begin{equation}
R_i=\sum_{t=0}^{T_i-1}r_{i,t},
\qquad
A_i=R_i-\frac{1}{G-1}\sum_{j\ne i}R_j.
\label{eq:call_advantage}
\end{equation}
}

We use \textbf{the same advantage $A_i$ to update both policies across all rounds} of rollout $i$, jointly improving semantic decisions and continuous execution for navigation success.

\subsection{Process and Outcome Rewards}
\label{sec:round_rewards}

For challenging long-horizon navigation tasks, terminal success alone provides sparse supervision across the many decisions and actions made over multiple interaction rounds.
We incorporate additional rewards derived from environment feedback at each interaction round. 
By distinguishing intermediate progress across rollouts, these rewards preserve relative learning signals even when no rollout reaches the goal, 
mitigating gradient degeneration when all rollouts fail:

{\small
\begin{equation}
r_t
=w_{\mathrm{succ}}r_t^{\mathrm{succ}}
+w_{\mathrm{arr}}r_t^{\mathrm{arr}}
+w_{\mathrm{prog}}r_t^{\mathrm{prog}}
+w_{\mathrm{info}}r_t^{\mathrm{info}}.
\label{eq:call_reward}
\end{equation}
}

\paragraph{Terminal  Success.}
We reward task success in the interaction round where it is confirmed by the environment's native success criterion:

{\small
\begin{equation}
r_t^{\mathrm{succ}}
=S_t,
\label{eq:terminal_reward}
\end{equation}
}
where $S_t \in \{0,1\}$ indicates whether task success is confirmed at round $t$ of rollout $i$.

\paragraph{First Arrival.}
We reward the agent once when it first enters the goal region, which provides an outcome signal that is easier to obtain than terminal success. Let $\eta^-_t$ and $\eta^+_t$ indicate whether the goal has ever been reached before and after a call, respectively:

{\small
\begin{equation}
r_t^{\mathrm{arr}}
=(1-\eta^-_t)\eta^+_t.
\label{eq:first_arrival_reward}
\end{equation}
}

\paragraph{Distance Progress.}
On challenging navigation tasks, both terminal success and first arrival can be difficult to achieve. We use process rewards to distinguish productive behavior even when neither outcome reward is obtained. We measure progress through a bounded proximity potential based on traversable distance $d(s)$ to the goal region. For each rollout $i$, we set the distance scale $D=\max(d(s_0),d_{\min})$ from its initial state $s_0$, using $d_{\min}>0$ as a lower bound. We then compute $r_t^{\mathrm{prog}}$ as the increase in goal proximity from the pre-call state $s^-_t$ to the post-call state $s^+_t$:

{\small
\begin{equation}
r_t^{\mathrm{prog}}
=\frac{D}{D+d(s^+_t)}
-\frac{D}{D+d(s^-_t)}.
\label{eq:progress_reward}
\end{equation}
}

\paragraph{Information Discovery.}
Distance progress alone overlooks active observation that reveals useful navigation cues without immediate movement toward the goal. We therefore reward newly discovered, depth-verified task-relevant entities. After each call, we count newly discovered entities $n_t$ by excluding the previously observed set $\mathcal S_t$, including the initial view, from the currently visible set $\mathcal V_t$. We add these discoveries to the pre-call cumulative count $c_t$ and reward its increase up to a total of $N_{\mathrm{info}}$ discoveries:

{\small
\begin{equation}
n_t
=|\mathcal V_t\setminus\mathcal S_t|,\qquad
r_t^{\mathrm{info}}
=\min(N_{\mathrm{info}},c_t+n_t)
-\min(N_{\mathrm{info}},c_t).
\label{eq:information_reward}
\end{equation}
}

\subsection{Joint Policy Optimization}

\paragraph{Tool Policy Optimization.}
We optimize the VLM brain policy $\pi_\theta$ using a REINFORCE objective~\cite{ahmadian2024back,li2023remax} with rollouts sampled from the current policy to improve movement decisions through semantic tool calls. 
We weight the mean log-probability of sampled response tokens by the trajectory-level advantage $A_i$:
{\small
\begin{equation}
\bar{\ell}^{\mathrm{text}}_{i,t}
=\frac{1}{|\mathcal{T}_{i,t}|}\sum_{n\in\mathcal{T}_{i,t}}
\log\pi_{\theta}(y_{i,t,n}\mid x_{i,t},y_{i,t,<n}),\qquad
\mathcal{L}_{\mathrm{text}}
=-\frac{\lambda_{\mathrm{text}}}{B}\sum_i\frac{1}{T_i}\sum_t
\operatorname{sg}(A_i)\bar{\ell}^{\mathrm{text}}_{i,t}.
\label{eq:text_objective_main}
\end{equation}
}
For each round, we average token log-probabilities over the sampled response positions $\mathcal{T}_{i,t}$. We weight this mean by $\operatorname{sg}(A_i)$, holding the trajectory advantage constant during backpropagation, then average over rounds within each rollout and the $B$ rollouts in the batch.

\paragraph{Action Policy Optimization.}
We optimize the action expert policy $\pi_\phi$ through policy gradients over sampled Flow-SDE paths, following Flow-GRPO and DPPO~\cite{liu2026flow,ren2025diffusion}. We clip the trajectory advantage to obtain $A_i^{\mathrm{exec}}=\operatorname{clip}(A_i,-a_{\max},a_{\max})$ using threshold $a_{\max}$ and use it to weight the path log-likelihood $\ell^{\mathrm{exec}}_{i,t,k}$ at execution cycle $k$:
{\small
\begin{equation}
\nabla_{Q,\phi}\mathcal{L}_{\mathrm{action}}
=-\mathbb{E}_{i,t,k}\!\left[
\operatorname{sg}(A_i^{\mathrm{exec}})
\nabla_{Q,\phi}\ell^{\mathrm{exec}}_{i,t,k}
\right].
\label{eq:action_gradient_main}
\end{equation}
}
We hold $A_i^{\mathrm{exec}}$ constant during backpropagation and average the weighted path gradients over cycles within each movement call, calls within each rollout, and then rollouts.

\paragraph{Joint Update with Separate Gradient Paths.}
\textbf{We couple the slow and fast policies through the shared trajectory-level advantage}, so that both semantic tool calls and action execution are optimized toward the overall navigation objective. Meanwhile, we maintain separate gradient paths:

{\small
\begin{equation}
g_\theta
=\nabla_\theta\mathcal L_{\mathrm{text}},
\qquad
g_{Q,\phi}
=\nabla_{Q,\phi}\mathcal L_{\mathrm{action}}.
\label{eq:gradient_routing}
\end{equation}
}
In this way, \textbf{joint RL coordinates both policies through shared navigation feedback while preventing action-policy gradients from directly altering the brain's tool-call distribution.}
We provide the complete algorithm in Algorithm~\ref{alg:joint_agentic_rl} and detailed optimization objectives in Appendices~\ref{sec:rl_algorithm}--\ref{sec:action_policy_details}.

\begin{figure}[!htb]
\begin{tcolorbox}[colback=white,colframe=black!55,boxrule=0.6pt,arc=0pt,
left=6pt,right=6pt,top=6pt,bottom=6pt]
\refstepcounter{vlnalgorithm}
\label{alg:joint_agentic_rl}
\noindent\textbf{Algorithm \thevlnalgorithm. Joint Agentic RL with Trajectory-Level Advantage Estimation}
\par\smallskip\hrule\smallskip
\small
\begin{tabular}{@{}r@{\hspace{0.7em}}p{\dimexpr\linewidth-2.5em\relax}@{}}
\multicolumn{2}{@{}p{\linewidth}@{}}{\textbf{Input}\quad Episode distribution $\mathcal D$, base brain $\pi_\theta$, initialized $(Q,\phi)$, frozen $E_\psi$, group size $G>1$, training steps $S$, round budget $T_{\max}$, text-loss weight $\lambda_{\mathrm{text}}$, action-advantage bound $a_{\max}$.}\\
\multicolumn{2}{@{}p{\linewidth}@{}}{\textbf{Output}\quad Updated $(\theta,Q,\phi)$.}\\
\multicolumn{2}{@{}p{\linewidth}@{}}{\textbf{Notation}\quad $U$: action chunk, $\xi$: Flow-SDE path, $o$: RGB, execution feedback, and trajectory trace.}\\[3pt]
1 & \textbf{for} $m=1,\ldots,S$ \textbf{do}\\
2 & \hspace*{1em}$\mathcal B\sim\mathcal D$ \hfill $\triangleright$ episode batch\\
3 & \hspace*{1em}\textbf{for each} $e\in\mathcal B$ \textbf{do}\\
4 & \hspace*{2em}\textbf{for} $i=1,\ldots,G$ \textbf{do}\\
5 & \hspace*{3em}$x_{i,0}\leftarrow\operatorname{Reset}(e)$\\
6 & \hspace*{3em}\textbf{for} $t=0,\ldots,T_{\max}-1$ until termination \textbf{do}\\
7 & \hspace*{4em}$y_{i,t}\sim\pi_\theta(\cdot\mid x_{i,t})$, retain response tokens\\
8 & \hspace*{4em}\textbf{if} $\operatorname{IsMovement}(y_{i,t})$ \textbf{then}\\
9 & \hspace*{5em}$Q_{i,t}\leftarrow[\operatorname{VLM}_\theta(\operatorname{Embed}(x_{i,t},y_{i,t})\oplus Q)]_Q$\\
10 & \hspace*{5em}$I^{\mathrm{ref}}_{i,t}\leftarrow\operatorname{RGB}(x_{i,t})$, $I_{i,t,0}\leftarrow I^{\mathrm{ref}}_{i,t}$, $k\leftarrow0$\\
11 & \hspace*{5em}\textbf{while} call active \textbf{do}\\
12 & \hspace*{6em}$C_{i,t,k}\leftarrow P_1(Q_{i,t})\oplus P_2(F_{i,t,k})$ \hfill Eq.~\eqref{eq:expert_condition}\\
13 & \hspace*{6em}$(U_{i,t,k},\xi_{i,t,k})\sim\operatorname{FlowSDE}_\phi(C_{i,t,k})$, retain $\xi_{i,t,k}$\\
14 & \hspace*{6em}$o\leftarrow\operatorname{Execute}(U_{i,t,k}[1\!:\!h])$, $I_{i,t,k+1}\leftarrow\operatorname{RGB}(o)$, $k\leftarrow k+1$\\
15 & \hspace*{5em}\textbf{end while}\\
16 & \hspace*{4em}\textbf{else} $o\leftarrow\operatorname{Execute}(y_{i,t})$ \hfill $\triangleright$ turn / stop\\
17 & \hspace*{4em}\textbf{end if}\\
18 & \hspace*{4em}$r_{i,t}\leftarrow w_{\mathrm{succ}}r^{\mathrm{succ}}_{i,t}+w_{\mathrm{arr}}r^{\mathrm{arr}}_{i,t}+w_{\mathrm{prog}}r^{\mathrm{prog}}_{i,t}+w_{\mathrm{info}}r^{\mathrm{info}}_{i,t}$, $x_{i,t+1}\leftarrow\operatorname{Append}(x_{i,t},y_{i,t},o)$\\
19 & \hspace*{3em}\textbf{end for}, $T_i\leftarrow$ number of completed rounds\\
20 & \hspace*{2em}\textbf{end for}\\
21 & \hspace*{2em}$R_i\leftarrow\sum_{t=0}^{T_i-1}r_{i,t}$, $A_i\leftarrow R_i-\frac{1}{G-1}\sum_{j\ne i}R_j$ \hfill $\forall i$\\
22 & \hspace*{1em}\textbf{end for}\\
23 & \hspace*{1em}$g_\theta\leftarrow\nabla_\theta\mathcal L_{\mathrm{text}}(\{A_i\})$ \hfill Eq.~\eqref{eq:text_objective_main}\\
24 & \hspace*{1em}$g_{Q,\phi}\leftarrow\nabla_{Q,\phi}\mathcal L_{\mathrm{action}}(\{A_i^{\mathrm{exec}}\},\theta\ \text{fixed})$ \hfill Eq.~\eqref{eq:action_gradient_main}\\
25 & \hspace*{1em}$(\theta,Q,\phi)\leftarrow\operatorname{OptimizerStep}(\theta,Q,\phi,g_\theta,g_{Q,\phi})$\\
26 & \textbf{end for}\\
\end{tabular}
\par\smallskip
\noindent $F_{i,t,k}$ fuses $E_\psi(I^{\mathrm{ref}}_{i,t})$ and $E_\psi(I_{i,t,k})$ (Section~\ref{sec:architecture}).
We use $A_i^{\mathrm{exec}}=\operatorname{clip}(A_i,-a_{\max},a_{\max})$ for the action objective. The action gradient passes through the VLM to $Q$. $\theta$ is updated only by $g_\theta$, and $\psi$ remains frozen.
\end{tcolorbox}

\end{figure}

\section{Experiments}
\label{sec:experiments}

\subsection{Experimental Setup}
\label{sec:implementation}

\paragraph{Datasets and Evaluation.}

\begin{figure}[!htb]
\centering
\includegraphics[width=0.6\textwidth]{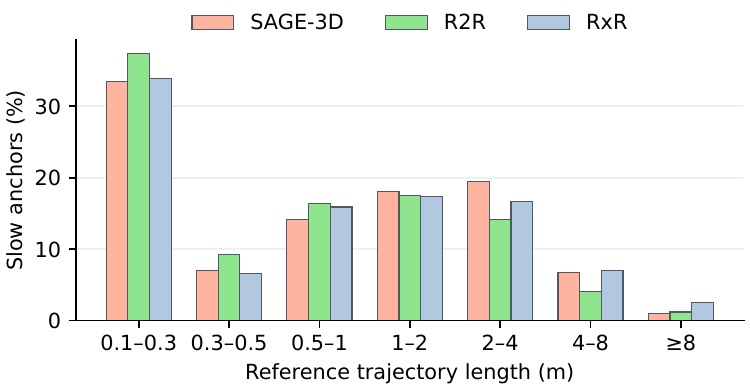}
\caption{\textbf{Reference Trajectory Lengths.} Across 1,000 sampled anchors per dataset, reference trajectories span multiple distance scales. The concentration of short trajectories partly reflects the 0.1\,m collection threshold.}
\label{fig:teacher_route_lengths}
\end{figure}

We evaluate \method{} on three mainstream VLN-CE benchmarks: SAGE-Bench~\cite{miao2025sage}, R2R-CE~\cite{anderson2018r2r,krantz2020beyond}, and RxR-CE~\cite{ku2020rxr}. 
We restrict navigation training to a shared pool of only 34,445 episodes, used for action expert adaptation with the VLM brain frozen and sampled at every RL training step. 
For  action expert adaptation, we generate request-grounded reference trajectories spanning multiple distance scales, as shown in Figure~\ref{fig:teacher_route_lengths}.

Following the official criteria, SAGE-Bench counts reaching the goal region as success, whereas R2R-CE and RxR-CE additionally require stopping there.
We report success rate (SR), success weighted by path length (SPL) to measure path efficiency, and, for R2R-CE and RxR-CE, navigation error (NE) as the final distance to the goal and oracle success rate (OSR) as success based on any goal-region visit. Full evaluations use the official evaluation sets with a 200-round limit per episode. During RL training, we additionally evaluate every 4 training steps on 100 fixed episodes per benchmark, each with a 100-round limit.

\paragraph{Model and Training Details.}
We pair a Qwen3.6-35B-A3B VLM brain~\cite{qwen36} with a 12-layer flow-matching action expert using the Qwen3.5-4B vision encoder~\cite{qwen35}. For expert adaptation, we train the contextual queries and action expert for approximately 2 epochs with the brain and visual encoder frozen, using a batch size of 64 request--trajectory pairs and a constant learning rate of $5\times10^{-5}$. We set the action prediction horizon $H=16$ and the execution prefix $h=4$.
For closed-loop RL, we jointly train the brain, contextual queries, and action expert for 200 training steps, keeping the visual encoder frozen. We use a constant learning rate of $10^{-6}$ and a global batch of 512 long-horizon rollouts per training step (8 rollouts for each of 64 episodes). 
Further details are provided in Appendices~\ref{sec:semantic_teacher_details} and~\ref{sec:training_details}.

\paragraph{Baselines.}
We compare \method{} with training-free tool use and representative SFT-based VLN methods. 
\textbf{Direct Tool Use} equips the base VLM brain with primitive movement, turning, and stopping tools.
For controlled comparisons, we train \textbf{NaVILA} and \textbf{DualVLN} on the same backbone and training episodes as \method{}. 
NaVILA fine-tunes the VLM to predict numerical motion commands, whereas DualVLN fine-tunes it to predict pixel goals for a continuous-action expert. 
We provide baseline implementation and training details in Appendices~\ref{sec:primitive_baseline} and~\ref{sec:retrained_baselines}.

\subsection{Main Results}
\label{sec:main_results}

\subsubsection{Learning to Navigate While Preserving General Capabilities}
\label{sec:zero_tradeoff}

\begin{table}[!htb]
\centering
\caption{\textbf{Navigation Results.} With matched backbones and training episodes, ZeroVLN achieves the highest SR across all three benchmarks without brain SFT. Higher OSR but lower SPL than NaVILA on R2R/RxR reveals a reachability--efficiency trade-off.}
\label{tab:zero_main}
\fontsize{8pt}{9pt}\selectfont
\setlength{\tabcolsep}{3pt}
\resizebox{\linewidth}{!}{%
\begin{tabular}{llccccccccccc}
\toprule
\multirow{2}{*}{\textbf{Method}} & \multirow[c]{2}{*}{\textbf{Training Episodes}} & \multirow{2}{*}{\shortstack{\textbf{Brain}\\\textbf{SFT}}} & \multicolumn{2}{c}{\textbf{SAGE-3D}} & \multicolumn{4}{c}{\textbf{R2R-CE}} & \multicolumn{4}{c}{\textbf{RxR-CE}} \\
\cmidrule(lr){4-5} \cmidrule(lr){6-9} \cmidrule(lr){10-13}
& & & \textbf{SR} $\boldsymbol{\uparrow}$ & \textbf{SPL} $\boldsymbol{\uparrow}$ & \textbf{NE} $\boldsymbol{\downarrow}$ & O\textbf{SR} $\boldsymbol{\uparrow}$ & \textbf{SR} $\boldsymbol{\uparrow}$ & \textbf{SPL} $\boldsymbol{\uparrow}$ & \textbf{NE} $\boldsymbol{\downarrow}$ & O\textbf{SR} $\boldsymbol{\uparrow}$ & \textbf{SR} $\boldsymbol{\uparrow}$ & \textbf{SPL} $\boldsymbol{\uparrow}$ \\
\midrule
\multicolumn{13}{l}{\bfseries\itshape Controlled Comparisons} \\
Direct Tool Use & 0 & $\times$ & \underline{39.5} & \underline{24.5} & 8.69 & 35.8 & 19.0 & 10.2 & 10.2 & 35.8 & 21.8 & 13.5 \\
DualVLN~\cite{wei2025dualvln} & 34.4K & $\checkmark$ & 21.2 & 18.0 & 8.82 & 39.0 & 15.4 & 11.2 & 10.8 & 41.1 & 17.7 & 9.99 \\
NaVILA~\cite{cheng2024navila} & 34.4K & $\checkmark$ & 23.5 & 22.1 & \underline{6.06} & \underline{51.8} & \underline{46.5} & \textbf{41.8} & \underline{6.65} & \underline{55.6} & \underline{48.7} & \textbf{43.0} \\
ZeroVLN (w/o RL) & 34.4K & $\times$ & 36.9 & 20.3 & 9.51 & 42.4 & 17.0 & 8.40 & 11.4 & 44.1 & 18.7 & 9.61 \\
\rowcolor{navilahighlight}
\textbf{ZeroVLN (Ours)} & \textbf{34.4K} & $\boldsymbol{\times}$ & \textbf{51.3} & \textbf{32.0} & \textbf{5.92} & \textbf{66.7} & \textbf{52.7} & \underline{29.2} & \textbf{6.44} & \textbf{64.6} & \textbf{51.4} & \underline{28.1} \\
\midrule
\multicolumn{13}{l}{\bfseries\itshape Published Results} \\
NaVid-SAGE~\cite{miao2025sage} & 500K & $\checkmark$ & 36.0 & 32.0 & -- & 42.0 & 31.0 & 29.0 & -- & -- & -- & -- \\
NaVILA-SAGE~\cite{miao2025sage} & 500K & $\checkmark$ & 46.0 & -- & -- & 51.0 & 38.0 & 36.0 & -- & -- & -- & -- \\
Uni-NaVid~\cite{zhang2024uninavid} & $\approx$80K & $\checkmark$ & -- & -- & 5.58 & 53.3 & 47.0 & 42.7 & 6.24 & 55.5 & 48.7 & 40.9 \\
StreamVLN~\cite{wei2025streamvln} & $\geq$150K & $\checkmark$ & -- & -- & 4.90 & 63.6 & 56.4 & 50.2 & 5.65 & -- & 54.4 & 45.4 \\
NavFoM (single view)~\cite{zhang2025navfom} & $\geq$182K & $\checkmark$ & -- & -- & 5.01 & 64.9 & 56.2 & 51.2 & 5.51 & -- & 57.4 & 49.4 \\
JanusVLN~\cite{zeng2026janusvln} & $\geq$169K & $\checkmark$ & -- & -- & 4.78 & 65.2 & 60.5 & 56.8 & 6.06 & -- & 56.2 & 47.5 \\
\bottomrule
\end{tabular}%
}
\end{table}

\textbf{\method achieves competitive navigation performance with substantially fewer training episodes.} 
As shown in Table~\ref{tab:zero_main}, \method learns navigation from 34,445 training episodes without dataset-provided expert trajectories or navigation-domain supervised fine-tuning (SFT) of the VLM brain. 
On SAGE-3D, it achieves 51.3\% SR, surpassing NaVILA-SAGE's 46.0\% with roughly 1/14 as many training episodes. 
On R2R and RxR, \method{} surpasses Uni-NaVid by 13.4\% and 9.1\% in OSR, respectively, with corresponding SR gains of 5.7\% and 2.7\%.
Together, these results demonstrate that \method{} offers a data-efficient alternative to navigation-specific brain adaptation while achieving competitive VLN-CE performance.

\textbf{\method{} outperforms SFT-based VLN frameworks with the same backbone and training data.}
DualVLN has to learn both geometric goal prediction and action execution~\cite{wei2025dualvln},
while NaVILA~\cite{cheng2024navila} benefits from a simpler training setup, as its predicted motion commands are executed directly by the simulator without requiring a learned action controller. 
This helps explain its consistently stronger performance than DualVLN across all benchmarks and metrics.
Despite also learning continuous action control like DualVLN, \method{} avoids requiring the VLM brain to learn a navigation-specific output interface. Instead, request-grounded supervision teaches the action expert to execute the brain's natural-language requests, while joint RL further aligns semantic decisions with continuous execution. Under the same backbone and training episodes, \method{} outperforms DualVLN by over 30\% in SR across all three benchmarks. Notably, it also surpasses NaVILA, whose motion commands are executed directly by the simulator without a learned action controller, by 27.8\%, 6.2\%, and 2.7\% on SAGE-3D, R2R, and RxR, respectively.

\textbf{The flexibility of semantic tool use comes with a trade-off in path efficiency.} On R2R and RxR, \method{} achieves higher OSR than NaVILA (66.7\% and 64.6\% vs. 51.8\% and 55.6\%), but lower SPL (29.2\% and 28.1\% vs. 41.8\% and 43.0\%). 
Rather than imitating dataset-provided expert trajectories, \method{} learns through interaction, 
allowing the agent to recover from blocked movement by revising its requests and exploring alternative paths. 
This flexibility improves goal reachability, although the additional exploration can lead to longer navigation paths and lower SPL.

\begin{figure}[!htb]\centering
\includegraphics[width=0.6\textwidth]{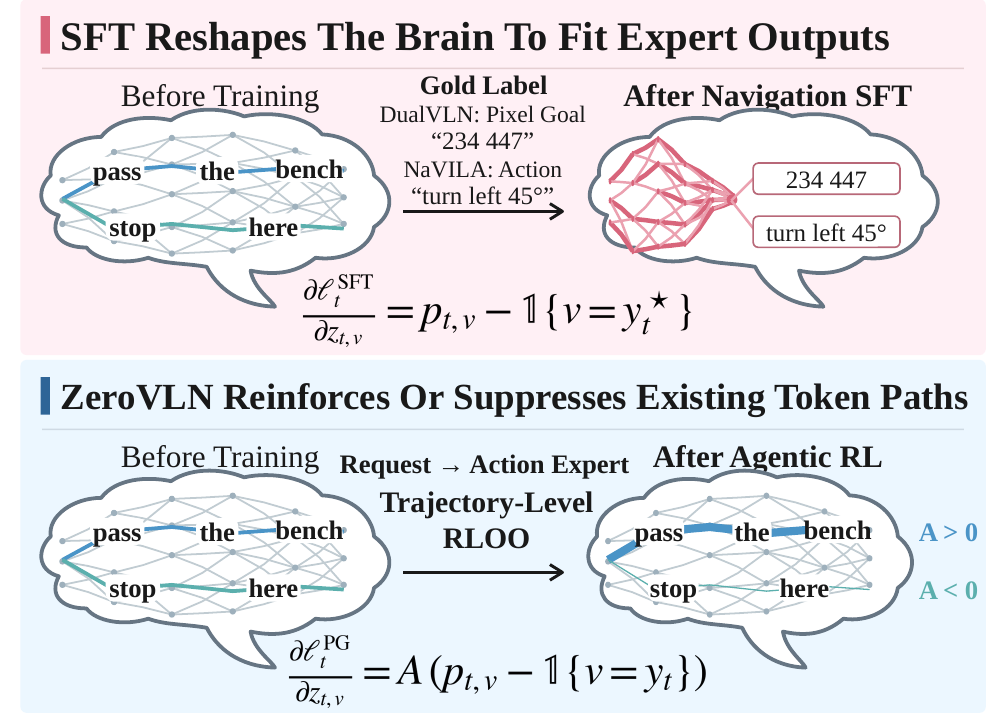}
\caption{\textbf{Navigation Training Mechanisms.} SFT fits the brain to prescribed navigation outputs. ZeroVLN instead reinforces its sampled semantic requests through navigation returns, building on the brain's existing capabilities.}
\label{fig:adapt_paths}
\end{figure}

\textbf{\method{} learns navigation while largely preserving the VLM brain's general capabilities.}
As illustrated in Figure~\ref{fig:adapt_paths}, Navigation SFT reshapes the brain's output distribution to fit navigation-specific targets, whereas our joint RL reinforces or suppresses the brain's own sampled tool-call token paths according to trajectory-level advantages.

\begin{figure}[!htb]\centering
\includegraphics[width=0.6\textwidth]{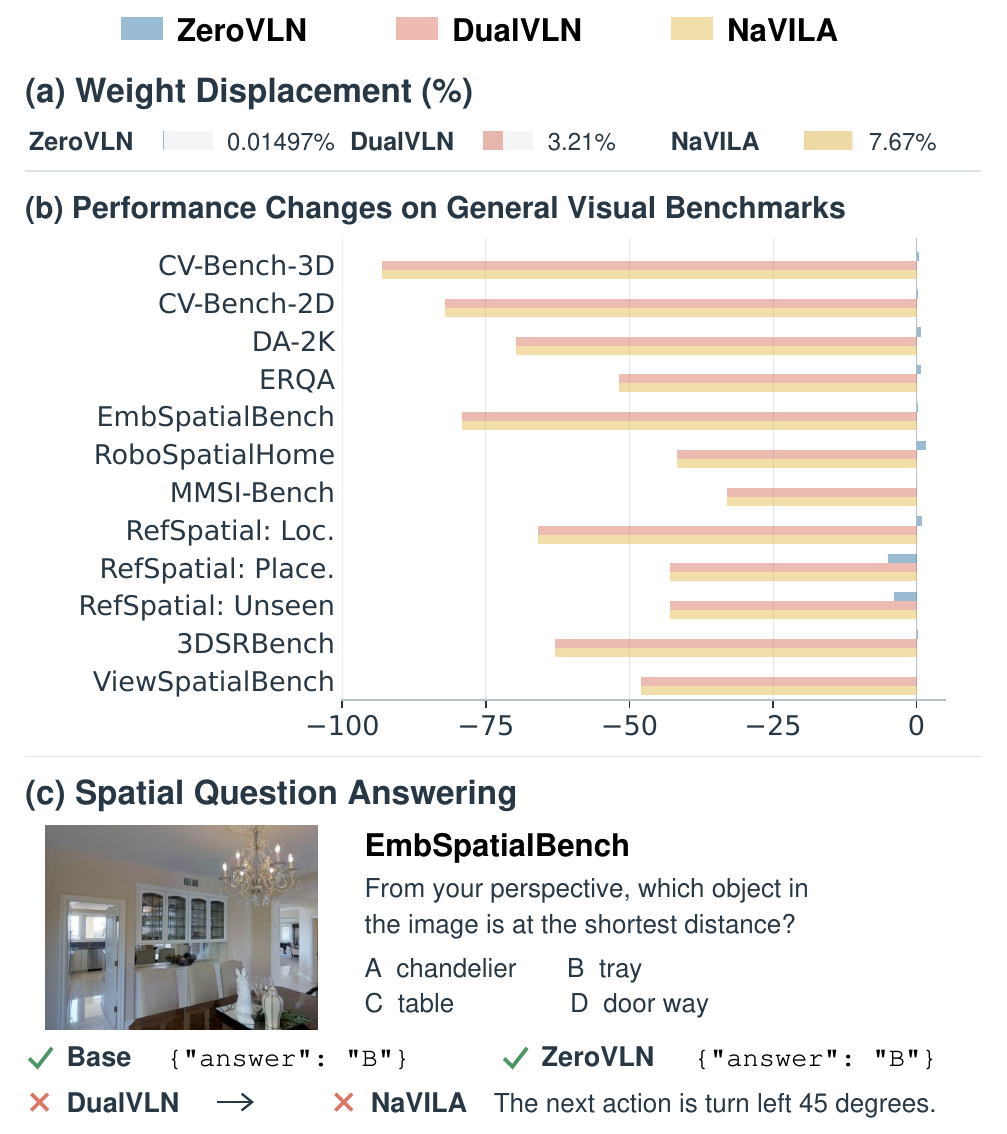}
\caption{\textbf{Small Brain Updates and Retained Capabilities.} ZeroVLN largely preserves general visual capabilities with substantially smaller VLM weight changes than NaVILA and DualVLN.}
\label{fig:zero_benefits}
\end{figure}

This distinction in training signals is reflected in the extent of VLM parameter changes.
As shown in Figure~\ref{fig:zero_benefits}, \method{} induces only 0.01497\% relative VLM weight displacement~\cite{ilharco2022editing,xuhong2018explicit}, compared with 3.21\% for DualVLN and 7.67\% for NaVILA.
This substantially smaller parameter shift is accompanied by stronger retention of general visual capabilities.
In a spatial reasoning example, both the base and ZeroVLN-tuned brains correctly identify the tray as the nearest object, whereas DualVLN and NaVILA generate navigation commands.
Such cases are not isolated. Across twelve general visual benchmarks, \method{} retains 59.16\% mean accuracy, close to the base brain's 59.47\%, while both SFT baselines score zero.
These qualitative and quantitative results jointly demonstrate that \method{} can acquire navigation capabilities with minimal changes to the VLM brain while largely preserving its general visual capabilities.
Further theoretical analysis and complete evaluation results are presented in Appendix~\ref{sec:brain_adaptation_theory} and ~\ref{sec:general_capability_results}, respectively.

\begin{figure}[!htb]
\centering
\includegraphics[width=0.85\textwidth]{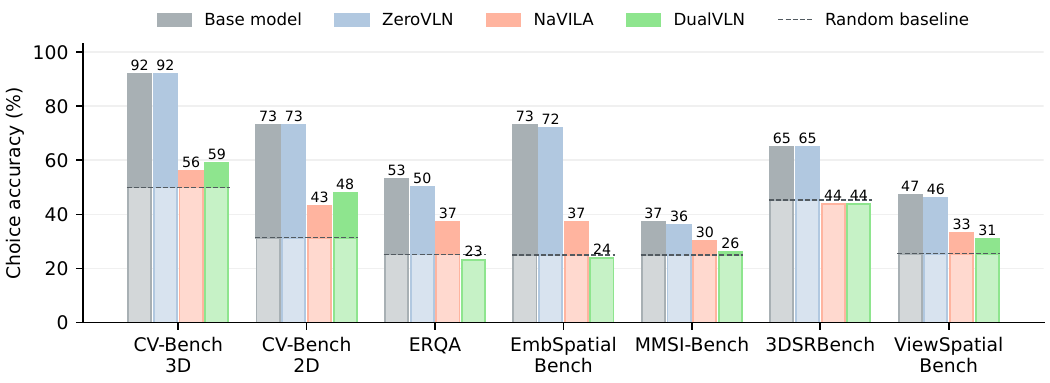}
\caption{\textbf{General Capabilities beyond Answer Formatting.} We read the probabilities of answer option tokens directly from the model's next-token vocabulary distribution, bypassing formatted answer generation. ZeroVLN largely retains base performance, whereas NaVILA and DualVLN perform close to chance on every benchmark.}
\label{fig:general7_choices}
\end{figure}

To test whether the zero scores reflect only failures to follow answer instructions, we sample 100 multiple-choice questions from each of seven benchmarks. After a fixed \texttt{Answer:} prefix, we read the probabilities of answer option tokens directly from the model's next-token vocabulary distribution. We select the most probable option without requiring the model to generate a formatted answer. As shown in Figure~\ref{fig:general7_choices}, NaVILA and DualVLN perform close to the random baseline on every benchmark. By contrast, \method{} achieves 62.00\% accuracy, close to the base brain's 62.86\%. \textbf{These results reveal severe degradation beyond instruction following: the general visual capabilities encoded in the models' parameters also deteriorate substantially on these tasks.}

\subsubsection{Why Does \method Work?}

\textbf{The Base Brain Already Makes Useful Navigation Decisions.}
As shown in Table~\ref{tab:zero_main}, Direct Tool Use achieves 39.5\% SR on SAGE-3D without navigation training, demonstrating that a general-purpose VLM can already leverage instructions and visual observations to make useful navigation decisions, which provides the foundation for our request-grounded supervision.

\begin{figure}[!htb]
\centering
\includegraphics[width=0.6\textwidth]{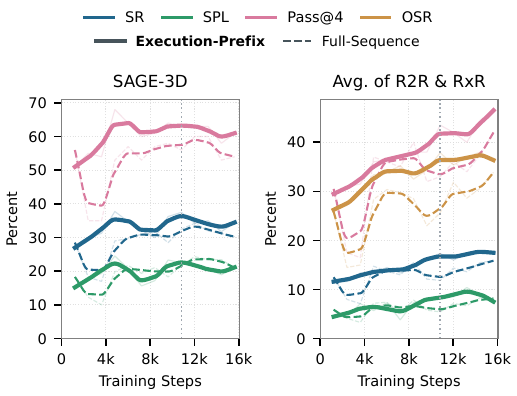}
\caption{\textbf{Execution-Prefix versus Full-Sequence Supervision.} With identical reference trajectories and a frozen brain, supervising the executed prefix yields stronger navigation performance and provides a useful initialization for joint RL.}
\label{fig:teacher_openloop}
\end{figure}

\textbf{Execution-Prefix Action Expert Adaptation Provides a Strong Initialization for RL.}
As shown in Figure~\ref{fig:teacher_openloop}, semantically aligned action adaptation raises SR from 27.0\% to 37.5\% on SAGE-3D and from 11.6\% to 17.4\% on average across R2R and RxR.
The supervision strategy also matters: on identical reference trajectories, execution-prefix supervision consistently outperforms full-sequence supervision, validating its better alignment with actual execution.
The adapted policy achieves a mean Pass@4 of 43.5\% on R2R/RxR, improving the likelihood of sampling successful rollouts.
This provides a stronger exploration policy for group-relative RL, mitigating degenerate advantage estimates under sparse terminal rewards.

\begin{figure}[!htb]
\centering
\includegraphics[width=0.85\textwidth]{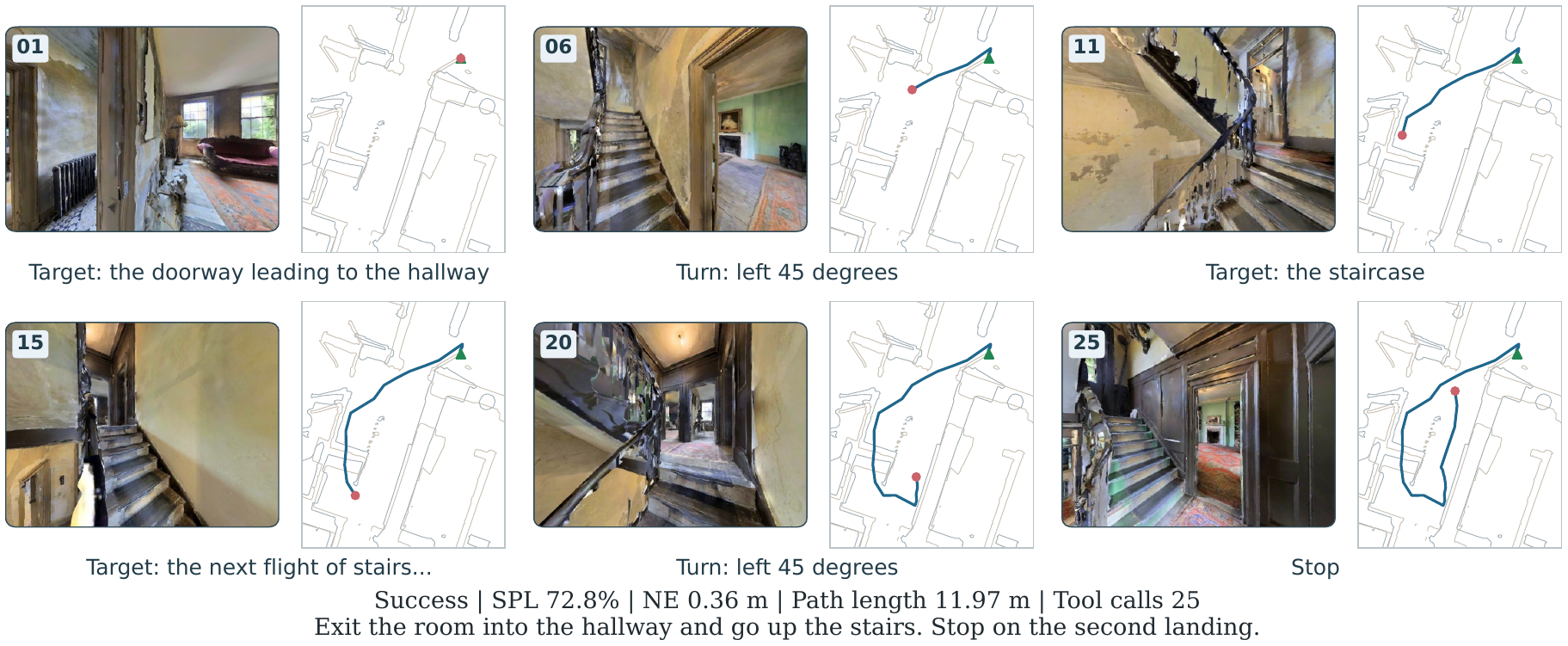}
\caption{\textbf{Recovery from Blocked Motion in R2R.} The brain changes its view and revises blocked climbing requests, reaches the second landing, and stops at round 25, illustrating feedback-driven coordination with the action expert.}
\label{fig:navigation_rollouts}
\label{fig:case_staircase}
\end{figure}

\textbf{Joint RL Translates Local Execution into End-to-End Navigation Success.} While supervised adaptation teaches the action expert to follow individual movement requests, successful navigation requires coordinating successive semantic decisions and continuous executions. As shown in Table~\ref{tab:zero_main}, joint RL raises SR from 17.0\% to 52.7\% on R2R, from 18.7\% to 51.4\% on RxR, and from 36.9\% to 51.3\% on SAGE-3D. The staircase example in Figure~\ref{fig:navigation_rollouts} illustrates this multi-round coordination. After encountering blocked motion, the brain changes its view and revises its movement request, which the action expert executes to reach the second landing before the brain successfully stops. This demonstrates how coordinated semantic decisions and continuous execution support recovery from local failures and successful completion of long-horizon navigation tasks.

\subsection{Ablation Studies}
\label{sec:ablation_studies}
\label{sec:roundab}
\label{sec:credit_comparison}

\begin{figure}[!htb]
\centering
\includegraphics[width=0.85\textwidth]{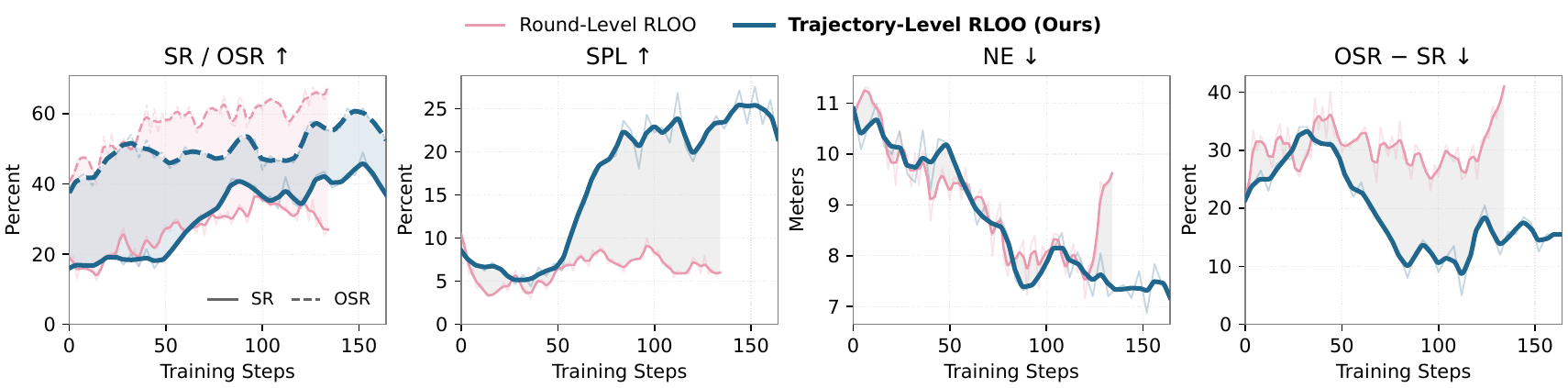}
\caption{\textbf{Round-Level versus Trajectory-Level RLOO.} Under matched rewards, trajectory-level credit improves SR and SPL, lowers NE, and narrows the OSR--SR gap, better connecting goal reaching with successful completion.}
\label{fig:credit_dynamics}
\end{figure}

\textbf{Trajectory-level advantages favor task completion and path efficiency, whereas round-level advantages emphasize goal reaching.} 
We compare the two credit assignment strategies under identical reward settings.
As shown in Figure~\ref{fig:credit_dynamics}, round-level advantages achieve higher OSR but exhibit declining SR and increasing NE during training, whereas trajectory-level advantages yield higher SR and SPL with lower NE, indicating more reliable task completion and better path efficiency despite fewer goal visits.
The key difference lies in how credit is assigned. 
Round-level advantages can reinforce a goal-reaching decision even if subsequent actions lead to failure, whereas trajectory-level advantages assign all rounds the same advantage derived from the complete navigation return. 
By accounting for both goal arrival and terminal success, trajectory-level credit favors decisions that contribute to successful task completion rather than isolated goal visits. 
Implementation details of the round-level variant are provided in Appendix~\ref{old:sec:credit_framework}.

\begin{figure}[!htb]
\centering
\includegraphics[width=0.85\textwidth]{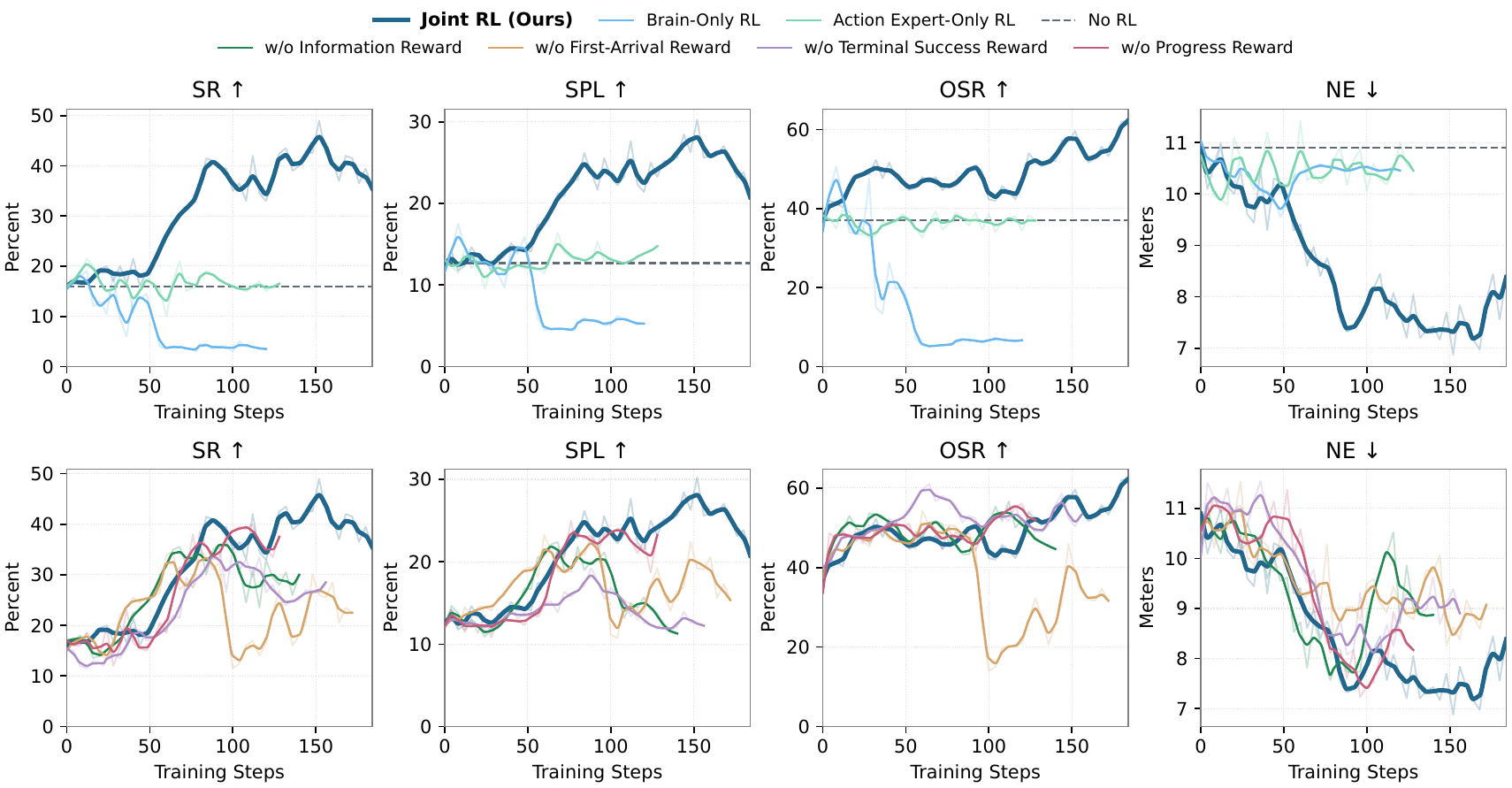}
\caption{\textbf{Joint Optimization and Reward Ablations.} Joint RL sustains improvements beyond updating either policy alone. Removing arrival, terminal-success, or process rewards weakens goal reaching, task completion, or training stability.}
\label{fig:ablation_overview}
\label{fig:d5_mean}
\label{fig:d5_information_reward}
\end{figure}

\begin{figure}[!htb]
\centering
\includegraphics[width=0.85\textwidth]{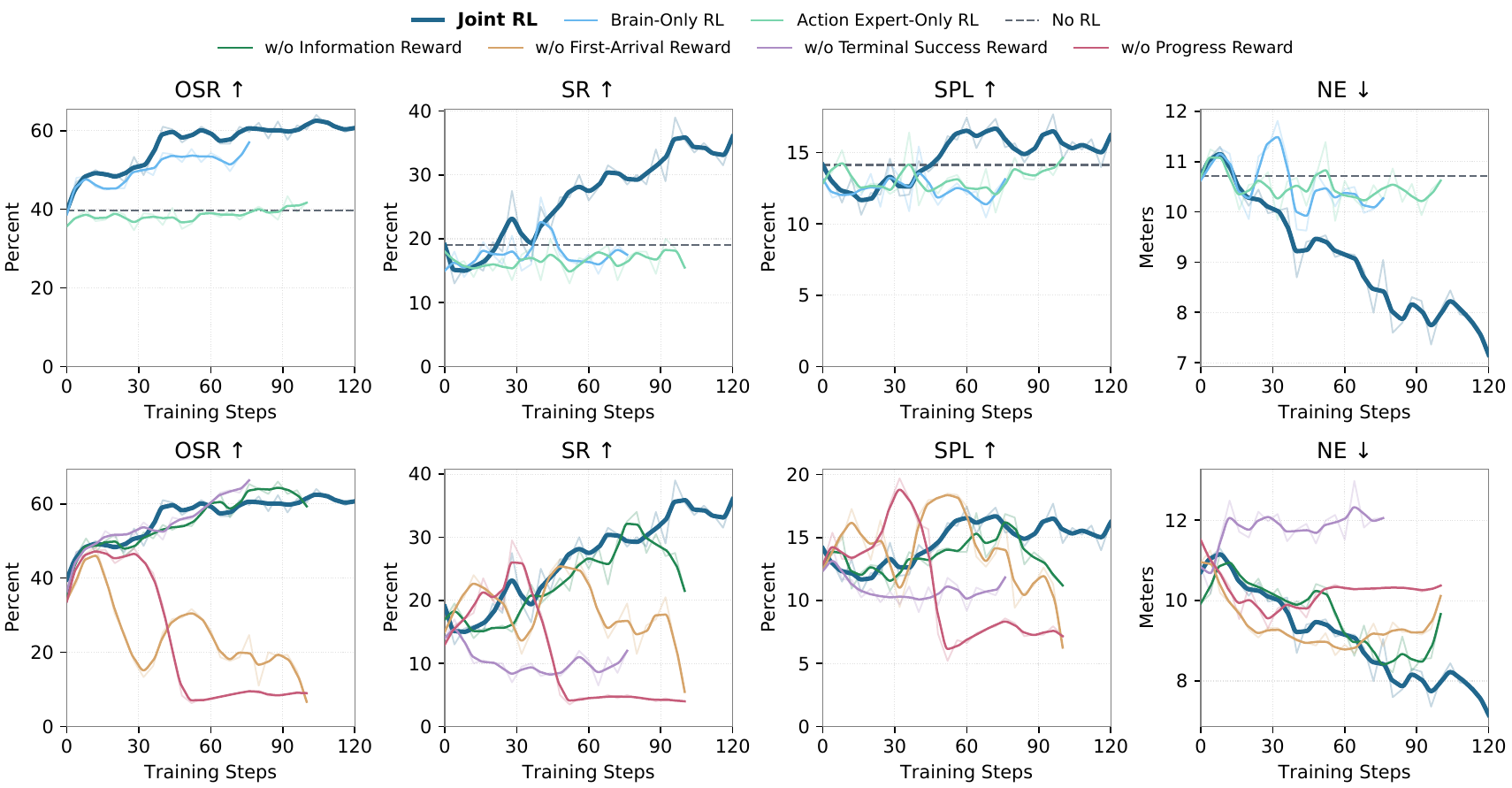}
\caption{\textbf{Joint Optimization and Rewards under Round-Level Credit.} We obtain conclusions consistent with those under trajectory-level credit (Figure~\ref{fig:ablation_overview}).}
\label{old:fig:ablation_overview}
\end{figure}

\textbf{Joint optimization enables coordinated adaptation of movement decisions and action execution.} As shown in Figure~\ref{fig:ablation_overview} (top), updating only the action expert yields limited improvements, suggesting that better execution alone cannot overcome the limitations of fixed semantic decisions. In contrast, updating only the brain initially improves navigation, but SR and OSR subsequently decline below initialization. This behavior is consistent with a growing mismatch between the brain's evolving movement requests and the fixed action expert. Joint RL, however, achieves sustained improvements across navigation metrics by allowing both policies to adapt together. These results highlight the importance of coordinated slow--fast optimization: improving either policy in isolation is insufficient when successful navigation depends on their continued alignment. Figure~\ref{old:fig:ablation_overview} (top) confirms this pattern under round-level credit: Joint RL improves SR and SPL beyond updating either policy alone.

\textbf{Each component of our reward design plays a significant role in sustaining stable navigation learning.}
We ablate each reward component under fixed trajectory-level credit assignment, as shown in Figure~\ref{fig:ablation_overview} (bottom).
\textbf{(1) Ablation of Outcome Rewards.} Removing the first-arrival reward severely degrades navigation performance across metrics, particularly OSR, demonstrating that the sparsity of terminal success rewards makes it difficult to learn effective goal-reaching behavior without denser intermediate feedback.
The first-arrival reward bridges this gap by providing a more immediate signal upon reaching the goal.
Conversely, without terminal success, OSR remains high while other metrics deteriorate, suggesting that rewarding arrival alone encourages goal visits rather than successful termination.
These results demonstrate that the two outcome rewards must work together to balance goal reaching and task completion.
\textbf{(2) Ablation of Process Rewards.} The process rewards further support learning when outcome rewards are scarce.
 Distance progress guides movement toward the goal, while information discovery encourages useful observation even without immediate spatial progress.
Removing either weakens or destabilizes training, demonstrating their complementary roles in sustaining navigation improvement.
We observe consistent trends under round-level credit assignment in Figure~\ref{old:fig:ablation_overview}, further supporting the complementary roles of goal-reaching and completion rewards.

\subsection{Consistency between Predicted Trajectories and Planner Goals}
\label{sec:teacher_goal_consistency}
\begin{figure}[!htb]
\centering
\includegraphics[width=0.5\textwidth]{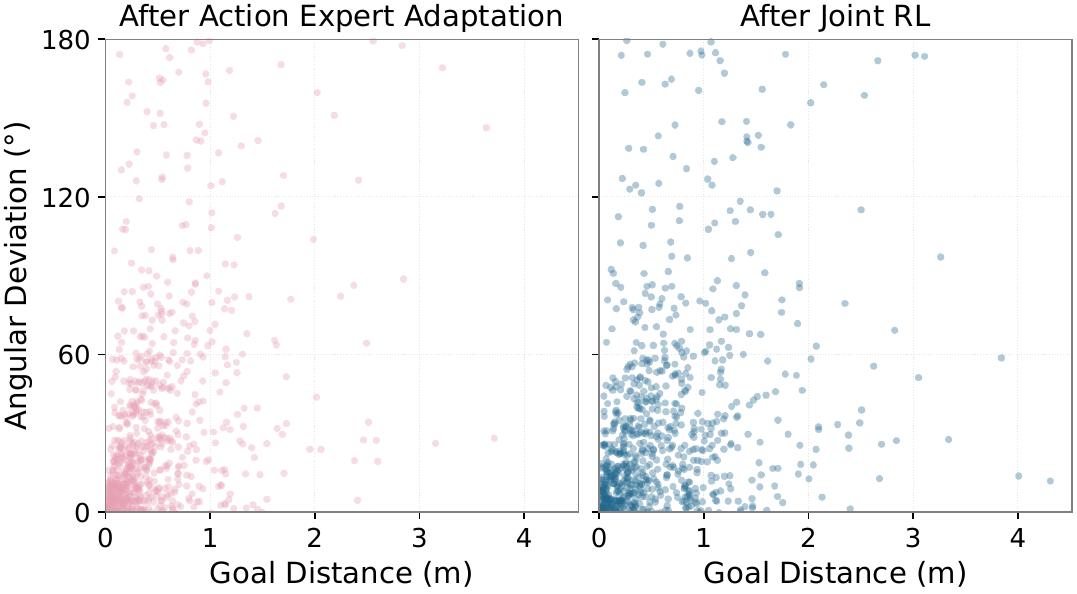}
\caption{Consistency between predicted trajectories and planner goals before and after joint RL, with  additional details  provided in Appendix~\ref{sec:goal_consistency_details}.}
\label{fig:teacher_goal_consistency}
\end{figure}

We assess the consistency between predicted trajectories and planner-grounded local goals before and after joint RL. Under identical observations and noise seeds, we measure the endpoint distances and angular deviations of predicted action sequences relative to the local goals. As shown in Figure~\ref{fig:teacher_goal_consistency}, both stages exhibit low endpoint distances and angular deviations, \textbf{demonstrating that supervised adaptation establishes goal-following behavior that is largely preserved after joint RL}.

\section{Conclusion}

We introduce \method{}, the first slow--fast VLN framework that enables a general-purpose VLM brain to guide action expert execution through semantic tool use, without navigation-domain SFT. 
We combine request-grounded and execution-prefix supervision with joint agentic RL, achieving strong navigation performance with limited data while largely preserving the VLM's general capabilities.
\textbf{Beyond navigation, this paradigm opens a path toward general-purpose embodied agents that acquire diverse physical skills through specialized action experts while retaining a shared VLM brain.}

\clearpage
\appendix
\renewcommand{\tabularxcolumn}[1]{m{#1}}
\raggedbottom
\addtocontents{toc}{\protect\setcounter{tocdepth}{2}}
\begingroup
\hypersetup{hidelinks}
\section*{Appendix}
\renewcommand{\contentsname}{Appendix Contents}
\tableofcontents
\endgroup
\clearpage
\section{Related Work}
\paragraph{\textbf{Slow--Fast Systems for Navigation.}}
Slow--fast architectures separate semantic planning from motion execution, allowing each policy to operate at its own timescale. Mobility VLA~\cite{chiang2024mobilityvla} uses a long-context VLM to select a goal frame from a demonstration tour and a topological navigation policy to reach it. NavGPT-2~\cite{zhou2024navgpt2} connects a frozen language model's vision-language representations to a navigation policy network. DualVLN~\cite{wei2025dualvln} trains its planner to predict pixel goals before learning contextual queries and a flow-matching action expert. ABot-N1~\cite{gong2026abotn1} similarly uses supervised reasoning and affordance pixels to guide a fast waypoint predictor. Beyond navigation, SmolVLA~\cite{shukor2025smolvla} decouples action prediction and execution through asynchronous inference, while Fast-in-Slow~\cite{chen2025fastinslow} partially shares parameters between slow reasoning and fast manipulation and coordinates them through dual-aware co-training. \textbf{\method{} retains this separation of planning and execution but connects the two policies through semantic requests, avoiding navigation-domain SFT for geometric goal prediction.}

\paragraph{\textbf{Navigation Planning with Existing Executors.}}
Navigation agents can use language-model reasoning to select actions executed by an external system. NavGPT~\cite{zhou2023navgpt} reasons over textual observations, navigation history, and candidate directions, while MapGPT~\cite{chen2024mapgpt} adds a linguistic map and adaptive multi-step planning. SayNav~\cite{rajvanshi2024saynav} grounds language-model plans in an incrementally constructed 3D scene graph and delegates their execution to an existing low-level navigation policy. AgenticNav~\cite{li2026agenticnav} exposes movement, depth, and memory as callable tools, with the harness converting selected image pixels into executable motion. NaVILA~\cite{cheng2024navila} instead predicts numerical movement commands in language. In the VLN-CE simulation setting considered here, these commands are executed by the system. Its legged-robot setting additionally uses a learned visual locomotion policy. ETP-R1~\cite{ye2025etp} applies closed-loop GRPO to topological planning while retaining the underlying controller. \textbf{Beyond selecting calls to an existing executor, \method{} learns the executor itself by treating the continuous-action head as a trainable tool.}

\paragraph{\textbf{Privileged Supervision for Navigation Action Policies.}}
Privileged geometry can supervise motion without requiring a human demonstration for every observation. For high-speed flight, Loquercio et al.~\cite{loquercio2021learning} train a sensorimotor policy in simulation by imitating a privileged expert, mapping onboard observations directly to collision-free trajectories. iPlanner~\cite{yang2023iplanner} backpropagates trajectory costs through a differentiable geometric costmap, while ViPlanner~\cite{roth2024viplanner} incorporates semantic costs to distinguish traversable surfaces and obstacles. NavDP~\cite{cai2025navdp} uses privileged simulation information to learn both trajectory generation and critic-based trajectory evaluation from local RGB-D observations. CAST~\cite{glossop2025cast} addresses instruction following by constructing counterfactual instruction--action labels from existing robot data. \textbf{In contrast, \method{} starts from the brain's actual movement requests and uses a privileged trajectory planner to generate matching reference trajectories.}

\paragraph{\textbf{Agentic RL and Credit Assignment.}}
DeepSeek-R1~\cite{deepseekai2025r1}, DAPO~\cite{yu2026dapo}, and GSPO~\cite{zheng2025group} improve language-model reasoning through RL. SAGE~\cite{huang2026does} targets reasoning efficiency, while R2M~\cite{huang2026real} aligns reward feedback with the evolving policy. Extending RL to multi-round agents requires connecting delayed outcomes to the decisions that produced them. ReTool~\cite{feng2026retool} interleaves language reasoning with code execution and uses outcome rewards to learn when and how to invoke tools. GiGPO~\cite{feng2026group} combines episode-level advantages with step-level comparisons at repeated environment states. PGPO~\cite{zheng2026pgpo} estimates state potentials from grouped rollout returns to distinguish intermediate actions. Agent Lightning~\cite{luo2025agent} decomposes agent trajectories into training transitions, while Agent-R1~\cite{cheng2025agent} separates generated outputs from environment feedback during policy optimization. VAGEN~\cite{wang2026vagen} introduces explicit world-model reasoning and bi-level credit assignment for visually grounded agents. In autonomous driving, DEFT-RLVR~\cite{huang2026deferred} combines verifiable trajectory selection with rubric-based reasoning rewards, assigning shared credit across its two-stage interaction. \textbf{\method{} adapts trajectory-level credit to navigation through complementary process and outcome rewards that account for exploration, goal reaching, and task completion.}

\paragraph{\textbf{Reinforcement Learning for Generative Action Policies.}}
Generative action policies require optimization through their action-generation process. Diffusion Policy Policy Optimization~\cite{ren2025diffusion} applies policy gradients to diffusion denoising steps for continuous control. Flow-GRPO~\cite{liu2026flow} converts deterministic flow sampling into stochastic paths to enable online group-relative optimization, providing a mechanism applicable to flow-based action generation. Flow Policy Gradients~\cite{yi2026flowpolicygradients} instead optimizes expressive flow policies without explicit likelihood computation for robot control. ConRFT~\cite{chen2025conrft} combines behavior cloning and Q-learning for offline VLA adaptation, followed by online consistency-policy optimization with human interventions. RIPT-VLA~\cite{tan2025riptvla} uses interactive rollouts and leave-one-out advantages to adapt VLA policies from sparse success rewards. \textbf{Building on action-policy RL, \method{} is, to our knowledge, the first to jointly optimize a VLM brain and a continuous-action expert through agentic RL in VLN-CE.}

\paragraph{\textbf{History Modeling and Spatial Planning.}}
Navigation policies use history to maintain spatial context beyond the current observation. HAMT~\cite{chen2021history} jointly encodes the instruction, past panoramic observations, and current view. DUET~\cite{chen2022think} combines local visual grounding with a global topological map for long-range planning. BEVBert~\cite{an2022bevbert} learns multimodal map representations that combine local metric structure with global topology, supporting spatial reasoning and long-range planning. GridMM~\cite{wang2023gridmm} organizes historical observations in an egocentric grid memory and aggregates instruction-relevant visual features. ETPNav~\cite{an2023etpnav} constructs an online topological map from predicted waypoints and couples cross-modal planning with an obstacle-avoiding controller. \textbf{\method{} likewise retains historical context, but supplies interaction records and the executed trajectory trace directly to the brain for semantic tool decisions.}

\section{Additional Details of Model Architecture and Training Paradigm}
\label{sec:method_details}

\subsection{Action Expert Architecture}
\label{sec:visual_conditioning}

Figure~\ref{fig:action_expert_architecture} illustrates our action expert. We provide all numerical settings and parameter counts in Appendix~\ref{sec:training_details}, including Tables~\ref{tab:action_expert_specs} and~\ref{tab:action_expert_parameters}. The architecture follows the dual-stream conditioning design of DualVLN~\cite{wei2025dualvln}. Our implementation uses a query-based visual pooler and a condition-fusion Transformer to combine the movement request with reference and current views.

\paragraph{Fusing Reference and Current Views.}
The frozen encoder $E_{\psi}$ extracts visual tokens from the reference image $I_t^{\mathrm{ref}}$, captured at the start of the movement call, and the refreshed image $I_{t,k}$. A shared linear projection maps both token sequences to the action expert's conditioning dimension. We add separate learned temporal embeddings to distinguish the reference and current views, concatenate the sequences, and apply a Transformer encoder. Learnable visual queries then cross-attend to the fused tokens, with a query residual connection and layer normalization, producing the fixed-length visual representation $F_{t,k}$. The reference image and semantic condition $Q_t$ remain fixed within the call, while current-view tokens are refreshed at each execution cycle.

\paragraph{Combining Semantic and Visual Conditions.}
The $N_q$ learnable queries have VLM hidden dimension $d_v$. Their final-layer states form the semantic condition $Q_t$ in Eq.~\eqref{eq:context_queries}. The operator $\oplus$ concatenates tokens, and $[\cdot]_Q$ selects the hidden states at the query positions.
In Eq.~\eqref{eq:expert_condition}, $P_1$ applies layer normalization, a linear projection, output normalization, and a learned semantic-channel embedding to $Q_t$. Since $F_{t,k}$ is already in the conditioning space, $P_2$ adds a learned visual-channel embedding without a further linear projection. Their concatenation $C_{t,k}$ forms the input to a condition-fusion Transformer within $f_{\phi}$, which mixes the semantic and visual tokens before velocity prediction. The input queries $Q$ are optimized separately from the action expert parameters $\phi$.

\paragraph{Generating Actions through Flow Matching.}
We linearly embed the noisy actions and add a learned position embedding for each of the $H$ action steps. A separate projection maps the fused conditions to the decoder width. A sinusoidal flow-time encoding followed by an MLP supplies a flow-time token, which is appended to the conditioning tokens to form the cross-attention memory. Each decoder block applies action self-attention, cross-attention to this memory, and a feedforward network, with pre-layer normalization and residual connections. A final layer normalization and linear projection predict a $d_a$-dimensional flow velocity at each action step. Starting from Gaussian noise, flow integration produces the normalized action sequence. Inverse normalization recovers displacement and heading increments for execution.

\paragraph{Action Execution and Feedback.}
Within each execution cycle, the predicted displacement and heading increments share the fixed local frame at the cycle's starting pose. We accumulate these increments into target poses and convert each target into a motion command using the agent's actual pose during execution. The coordinate frame is reset at the next cycle. Training targets follow the same convention, as detailed in Appendix~\ref{sec:action_target_details}. Each movement call returns the latest RGB observation obtained at the end of execution, together with executed motion and execution status. Reported cumulative motion is relative to the call's starting pose. The trajectory trace records the executed path and current heading and highlights blocked motion and collisions, providing spatial history for subsequent tool decisions.

\subsection{Request Grounding and Reference Trajectory Planning}
\label{sec:semantic_teacher_details}

\paragraph{Grounding Requests in Reachable Goals.}
As illustrated in Figure~\ref{fig:architecture} (I-a), our privileged semantic planner grounds the brain's movement request $q_t$ in a reachable local goal $g_t^\star$ using the agent state $s_t$, a mesh-derived semantic map, and training-only geometry. Dependency parsing extracts the target noun phrase, egocentric direction, and spatial relations from \tool{target}, \tool{where}, and the current movement stage of \tool{plan}. Lexical matching and a frozen text encoder retrieve matching object categories and mesh instances. We restrict the search to nearby instances supported by current or historical observations, then rank the candidates lexicographically by current depth-supported visibility, direction agreement, spatial-relation support, semantic match, and increasing distance to the object's bounding-box surface. These signals prioritize candidates rather than acting as independent rejection rules. We test candidates in this order and retain the first whose approach endpoint and reference route pass the geometric checks below.

For the current request $q_t$, we omit round indices on intermediate planner quantities. For each object instance $e$, we sample approach positions around its bounding-box boundary at increasing offsets from a preferred standoff distance $d_0$. We form the candidate set $\mathcal A(e)$ by removing occupied positions and positions unreachable from $s_t$. For each remaining position $g$, we compute the planned route length $L(g)$, test ground-plane line of sight to the object with the binary indicator $v_e(g)$, and measure the distance $d_e(g)$ to its bounding box. The selected approach position balances travel, geometric visibility, and proximity:
{\small
\begin{equation}
g(e)=\operatorname*{arg\,min}_{g\in\mathcal A(e)}
\left[L(g)+\lambda_v\bigl(1-v_e(g)\bigr)
+\lambda_d\max\bigl(0,d_e(g)-d_0\bigr)\right].
\label{eq:semantic_approach}
\end{equation}
}
The weights $\lambda_v$ and $\lambda_d$ control the visibility and distance penalties, respectively. The ground-plane visibility term is a soft geometric preference, distinct from the depth-supported camera visibility used in candidate ranking. An object is rejected if its selected endpoint exceeds the maximum allowed distance from its bounding box or its route violates the spatial constraints. Denoting the remaining geometrically verified instances by $\mathcal E$, the ranking and endpoint selection together anchor the request to the local goal:
{\small
\begin{equation}
e^\star=\operatorname*{lex\,arg\,max}_{e\in\mathcal E}
\bigl(\operatorname{Vis}(e),\operatorname{Dir}(e),\operatorname{Rel}(e),\operatorname{Match}(e),-\operatorname{Dist}(e)\bigr),
\qquad g_t^\star=g(e^\star).
\label{eq:semantic_route}
\end{equation}
}
The five ranking terms follow the visibility, direction, relation, semantic-match, and surface-distance criteria described above. Figure~\ref{fig:open_loop_training} illustrates how ``decorative painting'' is associated with its mesh instance and a nearby approach endpoint.

Doorway and room requests instead use mapped openings and their adjacent room identities to construct ordered waypoints before, within, and beyond the selected opening. Routes that cross unrelated rooms are rejected. If no object target is reachable but the request specifies an egocentric direction, the planner progressively shortens the displacement along that direction until it finds a collision-free endpoint without an unintended room crossing or a substantial directional detour. Requests without a feasible object, passage, or directional target are rejected.

\paragraph{Reference Trajectory Planning and Verification.}
For each candidate goal $g$, we search an occupancy grid whose traversable cells provide sufficient clearance for the agent's footprint. Diagonal transitions are allowed only when both neighboring cardinal cells are traversable, preventing corner cutting. To discourage routes that hug obstacles, we penalize edges whose minimum endpoint clearance $\operatorname{Clr}(a,b)$ falls below the preferred clearance $c_0$. A* search~\cite{hart1968formal} selects a grid path $P^\star(g)$ from the feasible paths $\mathcal P(g)$ connecting the position in $s_t$ to $g$ by minimizing
{\small
\begin{equation}
P^\star(g)=\operatorname*{arg\,min}_{P\in\mathcal P(g)}
\sum_{(a,b)\in P}\|b-a\|_2
\left[1+\lambda_c\left(\frac{\max(0,c_0-\operatorname{Clr}(a,b))}{c_0}\right)^2\right].
\label{eq:reference_path_cost}
\end{equation}
}
The weight $\lambda_c$ controls the clearance penalty. If the resulting path exceeds the allowed relative length increase over the ordinary shortest grid path, we use the latter to limit detours. The resulting route length supplies $L(g)$ in Eq.~\eqref{eq:semantic_approach}, coupling target selection to physical reachability and execution cost.

We simplify the selected route only when shortcuts remain collision-free and preserve its clearance cost, then apply spline smoothing and uniform arc-length resampling. Smoothing keeps both endpoints fixed and is progressively weakened if the curve exceeds the allowed increase in path length or reduction in minimum clearance, or violates the agent's footprint clearance. Every point and connecting segment must pass collision checks, otherwise the original safe polyline is retained. Stair requests use the native three-dimensional navigation mesh, with height changes constrained to the matched stair flight and requested ascending or descending direction. The verified geometry is converted into the reference trajectory $\rho_t:[0,L_t]\rightarrow\mathcal S$, parameterized by arc length with total length $L_t$.

Finally, we execute $\rho_t$ in the simulator and compare measured positions and headings with the planned waypoints. We reject trajectories with collisions, blocked motion, missing waypoint records, or position or heading tracking errors reaching their respective rejection thresholds. Acceptance additionally requires completion of the entire route and a matching terminal pose. The accepted trajectory and its originating request form the paired supervision used to train the action expert.

\paragraph{Selecting Requests for Verified Supervision.}
At each decision context, the brain samples up to $N_{\mathrm{cand}}$ candidate calls. The planner attempts grounding, trajectory planning, and execution verification, retaining the first successfully executed request. Each accepted call is grounded and planned once. Its context and verified reference trajectory form a slow-anchor training sample. Execution updates the observations before the brain generates the next call.

\subsection{Constructing Action Targets from Reference Trajectories}
\label{sec:action_target_details}

Each slow anchor pairs the originating navigation context and movement request with an execution-verified reference trajectory. As illustrated in Figure~\ref{fig:architecture} (I-b), resampling produces $K$ execution cycles, each with a condition $C_{t,k}$ and a normalized action target $U_{t,k}^\star$.

\paragraph{Resampling Trajectories into Execution Cycles.}
For a verified reference trajectory $\rho_t:[0,L_t]\rightarrow\mathcal{S}$ of arc length $L_t$, we sample $N+1$ reference poses, yielding $N$ action increments. Each pose contains a planar position and heading in the world frame. We fix $N$ to a multiple of the execution-prefix length $h$, producing $K=N/h$ execution cycles:
{\small
\begin{equation}
\Delta_t=L_t/N,\quad s_{t,j}^{\star}=\rho_t(j\Delta_t),\quad
s_{t,k}^{\star,\mathrm{fast}}=s_{t,(k-1)h}^{\star},\quad K=N/h.
\label{eq:route_resampling}
\end{equation}
}
Here $j=0,\ldots,N$ and $k=1,\ldots,K$. The reference pose $s_{t,k}^{\star,\mathrm{fast}}$ identifies the remaining trajectory for cycle $k$. It does not replace the agent's actual pose when constructing the targets. Scaling $\Delta_t$ with trajectory length provides finer increments for short requests and larger increments for long requests.

\paragraph{Action Targets in the Agent's Local Frame.}
For each cycle, we translate the remaining reference positions by the agent's actual starting position and rotate them by the negative of its actual heading. We unwrap reference headings and express them relative to this same starting heading. The first action increment runs from the actual starting pose to the first remaining reference pose. Subsequent increments are differences between successive reference poses, all expressed in the fixed coordinate frame of the cycle start. Each action comprises planar displacement and heading-increment components, with action dimension $d_a$. We retain the next $H$ increments and repeat the trajectory endpoint when fewer remain, giving zero motion after the final target pose. This construction accounts for deviations between the executed pose and the reference pose rather than assuming exact reference-trajectory replay.

\paragraph{Normalization and Flow-Matching Supervision.}
We apply fixed component-wise affine normalization to planar displacements and heading increments to obtain $U_{t,k}^{\star}\in\mathbb{R}^{H\times d_a}$, and invert this transform to recover physical actions before execution. Normalized targets are not clipped. Flow time $u$ is sampled from a logit-normal distribution, and Gaussian noise $\epsilon$ has the same shape as the complete target. The operator $[\cdot]_j$ selects action step $j$, and the expectation in Eq.~\eqref{eq:prefix_fm} covers training examples, flow times, and noise. Equation~\eqref{eq:prefix_fm} averages the prediction error over the $h d_a$ executed action components and the $K$ cycles of each slow anchor. All $H$ action tokens remain in the network input, while only the first $h$ positions contribute directly to the loss. Subsequent cycles supervise later motion with refreshed observations. The VLM parameters $\theta$ and encoder parameters $\psi$ remain frozen during this stage. Gradients update the input queries $Q$ and action expert parameters $\phi$.

\subsection{Joint Slow--Fast Policy Optimization}
\label{sec:rl_algorithm}
We detail the trajectory credit and gradient paths used by Algorithm~\ref{alg:joint_agentic_rl} in Section~\ref{sec:credit}.

\paragraph{Sharing Trajectory Returns across Decisions and Actions.}
\label{sec:credit_details}
Each interaction round comprises a tool call, its execution, and the resulting feedback. We sample $G$ independent rollouts from the same initial navigation episode and compute undiscounted returns over all their rounds. Equation~\eqref{eq:call_advantage} subtracts the other rollouts' mean return without standard-deviation normalization. All response tokens and executed action chunks in rollout $i$ receive its shared navigation credit.
The reward coefficients $w_{\mathrm{prog}}$, $w_{\mathrm{info}}$, $w_{\mathrm{arr}}$, and $w_{\mathrm{succ}}$ are nonnegative weights for the four terms in Section~\ref{sec:round_rewards}. Information discovery rewards at most $N_{\mathrm{info}}$ new entities, with total reward bounded by $w_{\mathrm{info}}N_{\mathrm{info}}$. The lower bound $d_{\min}$ prevents the initial-distance scale from vanishing. First-arrival flags persist after the agent leaves the goal region, so only the initial entry is rewarded. Rejected or otherwise invalid calls and STOP calls receive no first-arrival reward. Turning in place cannot trigger a first positional arrival.

\subsection{Tool Policy Optimization}
\label{sec:tool_policy_details}
For $B$ sampled rollouts, let $y_{i,t,n}$ denote sampled response token $n$ at round $t$ and $\mathcal{T}_{i,t}$ the corresponding token positions. The on-policy REINFORCE objective averages over sampled response tokens, rounds within each rollout, and the $B$ rollouts:
{\small
\begin{equation}
\begin{aligned}
\bar{\ell}^{\mathrm{text}}_{i,t}
&=\frac{1}{|\mathcal{T}_{i,t}|}\sum_{n\in\mathcal{T}_{i,t}}
\log\pi_{\theta}(y_{i,t,n}\mid x_{i,t},y_{i,t,<n}),\\
\mathcal{L}_{\mathrm{text}}
&=-\frac{\lambda_{\mathrm{text}}}{B}\sum_i\frac{1}{T_i}\sum_t
\operatorname{sg}(A_i)\bar{\ell}^{\mathrm{text}}_{i,t}.
\end{aligned}
\label{eq:text_objective}
\end{equation}
}
Here $\lambda_{\mathrm{text}}$ weights the text objective, and $\operatorname{sg}$ stops gradients through the advantage. This normalization controls the influence of response and episode length. The text branch uses an on-policy gradient without a PPO ratio.

\subsection{Action Policy Optimization}
\label{sec:action_policy_details}
The action expert policy $\pi_{\phi}$ generates an action chunk by starting from Gaussian noise and following a stochastic Flow-SDE path conditioned on $C_{i,t,k}$. Let $\ell^{\mathrm{exec}}_{i,t,k}$ denote the log-likelihood of the sampled generation path for execution cycle $k$. It sums the Gaussian transition log-probabilities over the denoising steps and all $H\times d_a$ action components. We clip the shared trajectory advantage from Eq.~\eqref{eq:call_advantage} to $A_i^{\mathrm{exec}}=\operatorname{clip}(A_i,-a_{\max},a_{\max})$ and assign it to every generation path in rollout $i$. For a single on-policy training step, the action-policy gradient is:
{\small
\begin{equation}
\nabla_{Q,\phi}\mathcal{L}_{\mathrm{action}}
=-\mathbb{E}_{i,t,k}\!\left[
\operatorname{sg}(A_i^{\mathrm{exec}})
\nabla_{Q,\phi}\ell^{\mathrm{exec}}_{i,t,k}
\right].
\label{eq:action_advantage_update}
\end{equation}
}
Positive action advantages reinforce sampled generation paths, while negative advantages suppress them. The operator $\operatorname{sg}$ stops gradients through the advantage. Only movement calls contribute: we average execution cycles within each call, calls within each rollout, and then rollouts. The action gradient updates $Q$ and $\phi$, propagating through the VLM to the input queries while holding the brain parameters $\theta$ fixed on this path. The generation path covers the full predicted chunk, although only its first $h$ physical actions are executed before visual replanning. We use Euler--Maruyama integration on a uniform flow-time grid from noise at $u=1$ to data at $u=0$. The drift uses the signed negative time step, while the Gaussian variance uses its absolute value. Flow-SDE sampling settings are provided in Appendix~\ref{sec:training_details}.

\subsection{Round-Level Advantage Estimation}
\label{old:sec:credit_framework}
Section~\ref{sec:credit_framework} assigns every call in rollout $i$ the same trajectory advantage $A_i$. We hold the group size $G$, rewards, optimizer, and gradient paths fixed. The round-level variant replaces $A_i$ with $A_{i,t}$ in both policy updates. At round $t$, it compares the return remaining in rollout $i$ with the mean remaining return of the other rollouts for the same episode. With discount factor $\gamma$, the remaining return and round-level advantage are:
{\small
\begin{equation}
R_{i,t}=\sum_{\tau=t}^{T_i-1}\gamma^{\tau-t}r_{i,\tau},\quad
A_{i,t}=R_{i,t}-\frac{\sum_{j\ne i}R_{j,t}}{G-1}.
\label{old:eq:call_advantage}
\end{equation}
}
A rollout that ends before round $t$ contributes zero to the baseline. The comparison is aligned by interaction round, rather than by visited state or navigation progress. Rewards collected before that round are excluded from its remaining return. We use the rewards in Section~\ref{sec:round_rewards} and the joint update in Section~\ref{sec:credit}. The tool policy uses $A_{i,t}$ directly, and the action expert uses $\operatorname{clip}(A_{i,t},-a_{\max},a_{\max})$.
Algorithm~\ref{alg:round_level_rloo} summarizes this variant using the same rollout collection and separate gradient paths as Algorithm~\ref{alg:joint_agentic_rl}.
\begin{tcolorbox}[colback=white,colframe=black!55,boxrule=0.6pt,arc=0pt,
left=6pt,right=6pt,top=6pt,bottom=6pt]
\refstepcounter{vlnalgorithm}
\label{alg:round_level_rloo}
\noindent\textbf{Algorithm \thevlnalgorithm. Joint Agentic RL with Round-Level Advantage Estimation}
\par\smallskip\hrule\smallskip
\small
\begin{tabular}{@{}r@{\hspace{0.7em}}p{\dimexpr\linewidth-2.5em\relax}@{}}
\multicolumn{2}{@{}p{\linewidth}@{}}{\textbf{Input}\quad $\mathcal D$, $\pi_\theta$, $(Q,\phi)$, frozen $E_\psi$, $G>1$, $S$, $T_{\max}$, $a_{\max}$, and $\lambda_{\mathrm{text}}$ as in Algorithm~\ref{alg:joint_agentic_rl}, discount $\gamma$.}\\
\multicolumn{2}{@{}p{\linewidth}@{}}{\textbf{Output}\quad Updated $(\theta,Q,\phi)$.}\\
\multicolumn{2}{@{}p{\linewidth}@{}}{\textbf{Notation}\quad $\mathcal R_e$: $G$ independent rollouts of episode $e$, $R_{i,t}$: return remaining at round $t$.}\\[3pt]
1 & \textbf{for} $m=1,\ldots,S$ \textbf{do}\\
2 & \hspace*{1em}$\mathcal B\sim\mathcal D$\\
3 & \hspace*{1em}\textbf{for each} $e\in\mathcal B$ \textbf{do}\\
4 & \hspace*{2em}$\mathcal R_e\leftarrow\operatorname{CollectRollouts}(e,\pi_\theta,Q,\phi,E_\psi,G,T_{\max})$ \hfill Algorithm~\ref{alg:joint_agentic_rl}, lines 4--20\\
5 & \hspace*{2em}$T\leftarrow\max_i T_i$, $R_{i,t}\leftarrow0$ \hfill $\forall i,\ 0\leq t\leq T$\\
6 & \hspace*{2em}\textbf{for} $i=1,\ldots,G$ \textbf{do}\\
7 & \hspace*{3em}\textbf{for} $t=T_i-1,\ldots,0$ \textbf{do}\\
8 & \hspace*{4em}$R_{i,t}\leftarrow r_{i,t}+\gamma R_{i,t+1}$\\
9 & \hspace*{3em}\textbf{end for}\\
10 & \hspace*{2em}\textbf{end for}\\
11 & \hspace*{2em}$A_{i,t}\leftarrow R_{i,t}-\frac{1}{G-1}\sum_{j\ne i}R_{j,t}$ \hfill $\forall i,\ t<T_i$\\
12 & \hspace*{2em}$A^{\mathrm{tool}}_{i,t}\leftarrow A_{i,t}$, $A^{\mathrm{exec}}_{i,t}\leftarrow\operatorname{clip}(A_{i,t},-a_{\max},a_{\max})$ \hfill $\forall i,\ t<T_i$\\
13 & \hspace*{1em}\textbf{end for}\\
14 & \hspace*{1em}$g_\theta\leftarrow\nabla_\theta\mathcal L_{\mathrm{text}}(A^{\mathrm{tool}})$ \hfill Eq.~\eqref{eq:text_objective}\\
15 & \hspace*{1em}$g_{Q,\phi}\leftarrow\nabla_{Q,\phi}\mathcal L_{\mathrm{action}}(A^{\mathrm{exec}},\theta\ \text{fixed})$ \hfill Eq.~\eqref{eq:action_advantage_update}\\
16 & \hspace*{1em}$(\theta,Q,\phi)\leftarrow\operatorname{OptimizerStep}(\theta,Q,\phi,g_\theta,g_{Q,\phi})$\\
17 & \textbf{end for}\\
\end{tabular}
\par\smallskip
\noindent $R_{j,t}=0$ for $t\geq T_j$. All $G-1$ other rollouts remain in the baseline.
Only $t<T_i$ contributes to updates. Action gradients use movement calls only.
\end{tcolorbox}

\section{Theoretical Analysis of Brain Adaptation and General Capability Retention}
\label{sec:brain_adaptation_theory}

Section~\ref{sec:zero_tradeoff} shows that \method{} learns navigation with substantially smaller VLM weight displacement than navigation SFT while largely retaining general visual capabilities. We analyze the mechanism behind this observation: \textbf{action expert adaptation makes the base brain's requests executable, while RL improves their use without requiring the brain to fit demonstration-defined outputs.} We first compare the resulting training signals, then examine their parameter requirements and the implications for general response stability.

\subsection{How SFT and RL Update the Brain}
\label{sec:adaptation_gradient_analysis}
Figure~\ref{fig:adapt_paths} contrasts fitting supervised outputs with reinforcing sampled requests. To formalize this distinction, we compare their response-level gradients under the same execution conditions. For a fixed navigation context and finite response set $\mathcal Y$, let $p_y=\exp(z_y)/\sum_v\exp(z_v)$ and $q_y$ denote the current and demonstration response probabilities. Fix the executor and continuation policy, giving each response $y$ an expected return $R_y$. Define $\mathcal L_{\mathrm{SFT}}=-\sum_yq_y\log p_y$ and $J=\sum_yp_yR_y$.

\paragraph{Proposition 1 (Gradients of SFT and RLOO).}
The SFT descent direction and expected, unnormalized RLOO ascent direction at response logit $z_y$ are
\begin{equation}
-\partial_{z_y}\mathcal L_{\mathrm{SFT}}=q_y-p_y,
\qquad
\partial_{z_y}J=p_y(R_y-J).
\label{eq:theory_logit_directions}
\end{equation}
\emph{Proof.} Using $\partial_{z_y}\log p_v=\mathbf1[v=y]-p_y$ gives
\begin{equation}
\begin{aligned}
-\partial_{z_y}\mathcal L_{\mathrm{SFT}}
&=\sum_vq_v(\mathbf1[v=y]-p_y)=q_y-p_y,\\
\partial_{z_y}J
&=\sum_vp_vR_v(\mathbf1[v=y]-p_y)=p_y(R_y-J).
\end{aligned}
\end{equation}
For independent on-policy samples $(y_i,R_i)$ satisfying $\mathbb E[R_i\mid y_i=y]=R_y$, the leave-one-out baseline $b_i$ is independent of the current sample. Since $\sum_v\nabla_zp_v=0$, its contribution cancels in expectation~\cite{ahmadian2024back}:
\begin{equation}
\begin{aligned}
\mathbb E[(R_i-b_i)\nabla_z\log p_{y_i}]
&=\sum_vR_v\nabla_zp_v-\mathbb E[b_i]\sum_v\nabla_zp_v
=\nabla_zJ.
\end{aligned}
\label{eq:theory_rloo_unbiased}
\end{equation}
\hfill$\square$

\textbf{RL selects responses by navigation value, whereas SFT also enforces a prescribed response distribution.} If all responses have equal expected return, the expected RL gradient vanishes, while SFT still changes the logits whenever $q\ne p$. This distinction explains the training mechanisms in Figure~\ref{fig:adapt_paths}. The semantic interface in Section~\ref{sec:architecture} allows useful language requests to be reinforced directly, without first converting the brain's outputs into geometric targets.

\paragraph{Return-Weighted Updates with Token and Round Averaging.}
The tool-policy objective in Section~\ref{sec:credit} also averages over tokens and rounds. For a rollout $\tau$ with $T$ rounds, let $h(\tau)=T^{-1}\sum_t\nabla_\theta\bar\ell_t^{\mathrm{text}}$ be its averaged token-score vector, as defined in Appendix~\ref{sec:tool_policy_details}. Conditional on the same initial episode and current policies, independent grouped rollouts with finite second moments give the expected negative weighted loss gradient
\begin{equation}
 g_{\mathrm{text}}
 =\lambda_{\mathrm{text}}\mathbb E[(R_i-b_i)h(\tau_i)]
 =\lambda_{\mathrm{text}}\mathbb E[(R-\mathbb E R)(h-\mathbb E h)].
\label{eq:theory_normalized_covariance}
\end{equation}
The identity follows from $\mathbb E[b_ih(\tau_i)]=\mathbb E[R]\mathbb E[h]$. Along any parameter-space direction $v$, the update is proportional to $\operatorname{Cov}(R,v^\top h)$. Thus the implemented objective retains the same principle: the brain changes according to how its sampled tool-call paths correlate with complete navigation returns.

\subsection{Why Navigation Can Require Smaller Brain Updates}
\label{sec:adaptation_kl_projection}
\label{sec:adaptation_parameter_analysis}
The gradient comparison identifies what drives adaptation. We next examine why fitting prescribed outputs can require additional parameter change, as suggested by the SFT comparisons in Section~\ref{sec:zero_tradeoff}. Improving navigation constrains response returns, whereas fitting demonstrations also constrains the response distribution. To compare these requirements at the same target return gain, consider a common fixed executor and the local approximation $p_{\theta_0+\Delta}=p_0+B\Delta$, with $B=\nabla_\theta p_{\theta_0}$ and base parameters $\theta_0$. For response-return vector $R$, the local return gradient is $g=B^\top R\ne0$. A target return gain $\delta>0$ requires $g^\top\Delta=\delta$. Assume that a demonstration distribution $q$ is locally realizable and provides the same gain, so $B\Delta=q-p_0$ is feasible and $R^\top(q-p_0)=\delta$.

\paragraph{Proposition 2 (Parameter Cost of Fitting Demonstrations).}
In this local model, the minimum-norm change achieving the return gain is $\Delta_{\mathrm{return}}=\delta g/\|g\|_2^2$. Any change $\Delta_{\mathrm{SFT}}$ that fits $q$ satisfies
{\small
\begin{equation}
\begin{aligned}
\Delta_{\mathrm{SFT}}&=\Delta_{\mathrm{return}}+w,\qquad g^\top w=0,\\
\|\Delta_{\mathrm{SFT}}\|_2^2
&=\frac{\delta^2}{\|g\|_2^2}+\|w\|_2^2
\geq\|\Delta_{\mathrm{return}}\|_2^2.
\end{aligned}
\label{eq:theory_parameter_cost}
\end{equation}
}
\emph{Proof.} Cauchy--Schwarz gives $\|\Delta\|_2\geq\delta/\|g\|_2$ for every return-feasible change, with equality at $\Delta_{\mathrm{return}}$. Demonstration fitting implies $g^\top\Delta_{\mathrm{SFT}}=R^\top B\Delta_{\mathrm{SFT}}=\delta$. Subtracting the return-only solution gives $g^\top w=0$, and the squared-norm identity follows by orthogonality.\hfill$\square$

\textbf{Demonstration fitting can add parameter change that does not improve navigation return.} The component $w$ matches the prescribed response distribution without increasing first-order return, and is nonzero whenever $q-p_0\ne B\Delta_{\mathrm{return}}$. This gives a parameter-space explanation for the distinction in Section~\ref{sec:zero_tradeoff}: NaVILA and DualVLN supervise particular motion commands or geometric targets, whereas \method{} can reward different semantic requests that lead to effective execution. The comparison complements the distributional perspective of RL's Razor~\cite{shenfeld2025rlrazor}. It characterizes local objective requirements, without assuming that RLOO attains the minimum-norm solution.

\paragraph{Action Expert Adaptation Reduces the Remaining Return Gap.}
The frozen-brain improvements in Figure~\ref{fig:teacher_openloop} show that navigation can improve by adapting execution alone. The request-grounded supervision in Section~\ref{sec:expert_adaptation} uses this route before updating the brain. Let $J_0$ be the base brain's return with the adapted action expert and $J_\star>J_0$ the target return. The remaining gap is $\delta=J_\star-J_0$. For an actual brain-update direction $d$ with $g^\top d>0$, define $\kappa=g^\top d/(\|g\|_2\|d\|_2)$. The local step along $d$ that closes this gap has size
{\small
\begin{equation}
\|\Delta_d\|_2=\frac{J_\star-J_0}{\|g\|_2\kappa}.
\label{eq:theory_residual_gap}
\end{equation}
}
At fixed local return sensitivity $\|g\|_2$ and positive update alignment $\kappa$, raising $J_0$ reduces the brain change required to reach $J_\star$. This connects the initialization gains in Figure~\ref{fig:teacher_openloop} to the small-displacement mechanism: \textbf{the action expert improves execution before RL adjusts the brain's requests, reducing the improvement left to brain adaptation.} During joint RL, the separate gradient paths in Eq.~\eqref{eq:gradient_routing} preserve this division of learning by updating the brain only through the return-weighted text objective. Together, an improved initialization and the absence of prescribed-output fitting explain how our training design can support navigation learning with small brain updates.

\subsection{Small Brain Updates and General-Response Stability}
\label{sec:adaptation_retention_bound}
Section~\ref{sec:zero_tradeoff} evaluates both weight displacement and general visual performance. To connect these observations, we bound how a parameter change affects the expected score of sampled general-purpose responses. The frozen-brain stage leaves this response distribution unchanged on identical inputs evaluated without navigation-specific modules. The following analysis concerns the subsequent brain update during joint RL.

For a fixed general-prompt distribution $\mu$, define $K(\theta)=\mathbb E_{x\sim\mu}D_{\mathrm{KL}}(\pi_\theta(\cdot\mid x)\|\pi_{\theta_0}(\cdot\mid x))$ and $V(\theta)=\mathbb E_{x\sim\mu,y\sim\pi_\theta}[a(x,y)]$, where $a\in[0,1]$ measures correctness or format validity. Let $\theta_1=\theta_0+\Delta$. Assume common response support, twice continuous differentiability with differentiation under expectation, and $\|\nabla^2K(\theta_0+t\Delta)\|_{\mathrm{op}}\leq M$ for $t\in[0,1]$.

\paragraph{Bounding Changes in General-Task Performance.}
Under these assumptions,
\begin{equation}
 K(\theta_1)\leq\frac{M}{2}\|\Delta\|_2^2,
 \qquad
 |V(\theta_1)-V(\theta_0)|
 \leq\sqrt{K(\theta_1)/2}
 \leq\frac{\sqrt M}{2}\|\Delta\|_2.
\label{eq:theory_retention_bound}
\end{equation}
\emph{Proof.} Since $K(\theta_0)=0$ and $\nabla K(\theta_0)=0$, Taylor's integral remainder gives
\begin{equation}
 K(\theta_1)=\int_0^1(1-t)\Delta^\top
 \nabla^2K(\theta_0+t\Delta)\Delta\,dt
 \leq\frac{M}{2}\|\Delta\|_2^2.
\end{equation}
For $a\in[0,1]$, the score difference is bounded by total variation. Applying Pinsker's inequality~\cite{canonne2023kltv} per prompt and Jensen's inequality over $\mu$ yields $|V(\theta_1)-V(\theta_0)|\leq\sqrt{K(\theta_1)/2}$.\hfill$\square$

\textbf{Smaller brain updates tighten the bound on general-response change when sensitivity is controlled.} Here $V(\theta)$ is the expected score under stochastic decoding, so the result applies to sampled responses. The bound depends on both displacement and curvature, whose value at initialization is the Fisher information~\cite{martens2020natural}. Relative displacement $\epsilon=\|\Delta\|_2/\|\theta_0\|_2$ enters through $\|\Delta\|_2=\epsilon\|\theta_0\|_2$. This connects the parameter comparison in Figure~\ref{fig:zero_benefits} to the capability analysis in Section~\ref{sec:zero_tradeoff}, while Table~\ref{tab:zero_general12} measures the retained behavior directly.

\textbf{Our central insight is to reduce what the brain must relearn to navigate.} The semantic interface preserves its native request space, action expert adaptation makes those requests executable, and joint RL improves their navigation value without matching prescribed geometric outputs. The analysis explains how this division can reduce the required brain change and, under bounded sensitivity, support stable general responses. It provides a theoretical account of the main-text finding that \method{} acquires navigation skills while largely retaining the base brain's general capabilities.

\section{Experimental Setup and Baselines}
\label{sec:additional_experiments}
\subsection{Training Data and Optimization Settings}
\label{sec:training_details}

\paragraph{Action Expert Architecture and Parameter Counts.}
Figure~\ref{fig:action_expert_architecture} and Tables~\ref{tab:action_expert_specs} and~\ref{tab:action_expert_parameters} summarize the implementation of the architecture in Appendix~\ref{sec:visual_conditioning}. The semantic and visual branches provide 16 and 32 tokens, respectively. The flow-time MLP adds one token to their concatenation, forming a 49-token cross-attention memory. Flow integration uses three steps. Each movement call permits up to four execution cycles. The 34.784M trainable action expert parameters comprise 0.792M in the semantic adapter, 3.365M in the visual adapter and pooler, 1.774M in condition fusion, 0.000768M in channel embeddings, and 28.852M in the velocity network. We load only the vision tower from Qwen3.5-4B. Open-loop adaptation updates the contextual queries and action expert with the brain and visual encoder frozen. Joint RL additionally updates the brain through the text objective.

\begin{table}[!htbp]
\centering\small
\caption{\textbf{ZeroVLN Action Expert Architecture.} We combine semantic queries with visual features to condition a flow matching decoder, then execute the first four of its 16 predicted action steps before visual replanning.}
\label{tab:action_expert_specs}
\newcommand{\expertspecrows}[1]{%
\begin{tabular}[b]{@{}>{\raggedright\arraybackslash}m{0.15\textwidth} >{\raggedright\arraybackslash}m{\dimexpr\linewidth-0.15\textwidth-2\tabcolsep\relax}@{}}
#1
\end{tabular}}
\begin{tabularx}{0.85\linewidth}{@{}>{\raggedright\arraybackslash}m{0.14\linewidth} >{\raggedright\arraybackslash}X@{}}
\toprule
\textbf{Component} & \expertspecrows{\textbf{Setting} & \textbf{Value}} \\
\midrule
Semantic Interface & \expertspecrows{
Input & Natural-Language Movement Request \\
Contextual Queries & 16 Queries Of Dimension 2,048 \\
} \\
\midrule
Visual Branch & \expertspecrows{
Encoder & Qwen3.5-4B Vision Tower (Frozen) \\
Feature Dimension & 2,560 \\
Temporal Fusion & Single-Layer Transformer \\
Token Pooling & Cross-Attention With 32 Queries \\
} \\
\midrule
Condition Fusion & \expertspecrows{
Fusion Module & Single-Layer Transformer \\
Output & 48 Tokens Of Dimension 384 \\
} \\
\midrule
Flow-Matching Decoder & \expertspecrows{
Architecture & Transformer Decoder With 12 Layers \\
Hidden Dimension & 384 \\
Attention Heads & 6 \\
Feedforward Dimension & 1,536 \\
Flow-Time Embedding & Two-Layer MLP \\
Cross-Attention Memory & 49 Tokens \\
} \\
\midrule
Action Output & \expertspecrows{
Action Dimension & 3 \\
Prediction Horizon & 16 Action Steps \\
Executed Prefix & 4 Action Steps Per Cycle \\
} \\
\bottomrule
\end{tabularx}
\end{table}

\begin{table}[!htbp]
\centering
\caption{\textbf{ZeroVLN Action Expert Parameter Counts.} Parameter counts exclude the VLM brain.}
\label{tab:action_expert_parameters}
\begin{tabular*}{0.6\linewidth}{@{}l@{\extracolsep{\fill}}ll@{}}
\toprule
\textbf{Parameter Group} & \textbf{Update Status} & \textbf{Parameters (M)} \\
\midrule
Action Expert & Trainable & 34.8 \\
Contextual Queries & Trainable & 0.033 \\
Visual Encoder & Frozen & 333.5 \\
\midrule
Trainable Parameters & - & 34.8 \\
\textbf{Total Parameters} & - & \textbf{368.3} \\
\bottomrule
\end{tabular*}
\end{table}

\paragraph{Request-Grounded Training Data.}
Table~\ref{tab:teacher_data} gives the episode and anchor counts for each dataset. We count each distinct scene, episode ID, and instruction as one source episode. The brain generates two candidate calls per decision and samples two more ($N_{\mathrm{cand}}=4$ in total) if neither passes the verification procedure in Appendix~\ref{sec:semantic_teacher_details}. We retain execution-verified movement calls with four complete execution cycles and at least 0.1\,m of accumulated translation. The planner records 65 poses along each reference trajectory. Keeping every fourth pose gives 17 poses and 16 action increments, divided into four execution cycles. \method{} (w/o RL) denotes the action-expert-adapted checkpoint used to initialize RL and serves as the No RL baseline in the update-scope ablation.
\begin{table}[!htbp]
\centering
\caption{\textbf{Training Data Collected With The Trajectory Planner.} Retention is the percentage of collected anchors passing the length and execution-validity filters. Each retained anchor contains four execution cycles.}
\label{tab:teacher_data}
\begin{tabularx}{0.6\linewidth}{@{}>{\raggedright\arraybackslash}X>{\raggedright\arraybackslash}X>{\raggedright\arraybackslash}X>{\raggedright\arraybackslash}X@{}}
\toprule
\textbf{Dataset} & \textbf{Episodes} & \textbf{Anchors} & \textbf{Retained (\%)} \\
\midrule
SAGE-3D & 14,628 & 108,860 & 61.1 \\
R2R & 10,069 & 140,610 & 67.3 \\
RxR & 9,748 & 135,366 & 67.9 \\
\midrule
Total & 34,445 & 384,836 & 65.6 \\
\bottomrule
\end{tabularx}
\end{table}

\paragraph{Trajectory Planner Settings.}
For the planner in Appendix~\ref{sec:semantic_teacher_details}, we use a preferred standoff distance $d_0=0.3\,\mathrm{m}$, visibility weight $\lambda_v=1$, and distance weight $\lambda_d=2$, with distances measured in meters. Selected endpoints must lie within $0.9\,\mathrm{m}$ of the object's bounding box. A* searches an eight-connected occupancy grid with preferred clearance $c_0=0.6\,\mathrm{m}$ and clearance weight $\lambda_c=0.8$. We fall back to the ordinary shortest grid path if the clearance-aware route is more than $25\%$ longer. Smoothing is weakened if it increases path length by more than $5\%$ or reduces minimum clearance by more than $0.05\,\mathrm{m}$. Execution verification rejects position errors reaching $0.035\,\mathrm{m}$ or heading errors reaching $2^\circ$.

\paragraph{Open-Loop Adaptation and Joint RL Settings.}
Both training stages use AdamW with $\beta_1=0.9$, $\beta_2=0.999$, numerical epsilon $10^{-8}$, and no learning-rate warmup. Weight decay is 0.01 during open-loop adaptation and zero during RL. Open-loop training runs on four 16-device nodes, with one slow anchor per device per training step, a global batch of 64 anchors, and gradient clipping at 1. Each anchor contains four execution cycles. We sample SAGE-3D, R2R, and RxR at a ratio of 1:1:2 in open-loop training and optimize the queries and action expert at a constant learning rate of $5\times10^{-5}$. For flow-matching supervision, the logit-normal flow-time distribution uses an underlying normal distribution with mean 0 and standard deviation 1. Each epoch contains 6,014 training steps. The RL initialization follows 10,826 training steps, or approximately 1.8 epochs. RL uses eight 16-device nodes and samples the datasets at a ratio of 2:1:1. Each training step contains 64 episodes with eight rollouts each, totaling 512 rollouts. The brain, queries, and action expert use a constant learning rate of $10^{-6}$, while the fast visual backbone remains frozen. The RL training budget is 200 training steps. We set the return discount $\gamma=1$ for both credit estimators, the text-loss weight $\lambda_{\mathrm{text}}=0.1$, and the action-advantage clipping threshold $a_{\max}=1$. Distance progress, first arrival, and terminal success each have weight 1. We set $N_{\mathrm{info}}=5$ and $w_{\mathrm{info}}=0.01$, capping the information reward at 0.05 per episode. Distance progress uses $d_{\min}=1\,\mathrm{m}$. The action sampler uses three uniform Flow-SDE transitions from $u=1$ through $u=2/3$ and $u=1/3$ to $u=0$, with noise scale 0.08. The drift step is $-1/3$, and the Gaussian variance uses its absolute value. The numerical variance floor is $10^{-6}$. The main single-training-step action objective is evaluated at the rollout policy, as in Eq.~\eqref{eq:action_advantage_update}.

\paragraph{Context Management.}
The brain has a 131,072-token context limit, with thinking disabled throughout collection, RL, and navigation evaluation. Supervision collection uses temperature 1.0 and top-$p$ 0.95, with a 256-token text budget. RL sampling uses temperature 1.0, top-$p$ 1.0, no top-$k$ truncation or repetition, presence, or frequency penalties, and a 512-token completion limit per round. Navigation evaluation uses temperature 0.7 and top-$p$ 0.95, with a 256-token text budget excluding reserved query tokens. Action expert data collection and RL allow 100 and 50 rounds per episode, respectively. On SAGE-3D, the navigation camera provides 640$\times$480 RGB observations with a horizontal field of view of $115^\circ$ at a height of 1.2\,m. We append tool calls, execution feedback, and observations to the interaction history. Only the final image of a movement call is appended as the next slow observation, separately from the tool result that reports motion and execution status. Intermediate images condition fast execution. During long evaluation episodes, history is compressed only when the input, output budget, and an 8,192-token reserve exceed the service window. Compression retains the initial task and recent eight rounds, samples one complete round from every four older rounds, and removes older past-trajectory snapshots while preserving the latest cumulative trajectory. Retained calls remain paired with their feedback and observations. Original round indices identify gaps in the sampled history.

\paragraph{Comparing Execution-Prefix and Full-Sequence Supervision.}
Figure~\ref{fig:teacher_openloop} compares Execution-Prefix Supervision and Full-Sequence Supervision on identical reference trajectories with the brain frozen. Both variants use 100 fixed episodes per dataset and four rollouts per episode, each with a 100-round budget. SR averages the outcomes of these rollouts. Pass@4 counts an episode as successful when at least one of its four rollouts succeeds. With the data, initialization, training settings, and evaluation held fixed, only the supervised action positions change: both variants predict 16 actions and execute four, while the flow-matching loss covers either the first four or all 16 actions (Section~\ref{sec:expert_adaptation}). Figure~\ref{fig:teacher_route_lengths} uses 1,000 sampled training anchors per dataset. The high frequency of short reference trajectories reflects the collection threshold: over half of the trajectories in the 0.1--0.3\,m bin lie within 1\,cm of 0.1\,m. Lengths up to 2\,mm below this cutoff are included in the first bin.

\paragraph{Evaluating Credit Assignment, Joint Updates, and Rewards.}
\label{sec:rl_ablation_protocol}
The upper row of Figure~\ref{fig:ablation_overview} compares update scopes using trajectory-level credit. The lower row compares reward ablations under the same trajectory-level credit assignment. Figure~\ref{old:fig:ablation_overview} additionally reports update-scope and reward ablations using round-level advantage estimation. All variants start from the same action-expert-adapted checkpoint. Each evaluation uses one rollout on each of 100 fixed episodes per dataset and a 100-round budget. In Figure~\ref{fig:ablation_overview}, SR and NE are averaged equally over R2R and RxR, while SPL is averaged equally over SAGE-3D, R2R, and RxR. Its OSR is the equal-weight mean of SAGE-3D arrival success and R2R/RxR OSR. The direct credit-assignment comparison in Section~\ref{sec:credit_comparison} holds rewards and update settings fixed and compares round-level and trajectory-level advantage estimation. In Figure~\ref{fig:credit_dynamics}, all four metrics (SR, OSR, SPL, and NE) are averaged equally over R2R and RxR, and the OSR--SR gap is computed from these same averages. Together, these three ablations test how credit coordinates decisions across rounds, whether both policies need to adapt, and which reward signals sustain navigation learning.

\subsection{Why Supervision Should Follow the Brain's Requests}
\label{sec:anchoring_alternatives}
We examine why the action expert should learn from the brain's actual requests, as proposed in Section~\ref{sec:expert_adaptation}. The alternatives expose two distinct mismatches: constructed requests can differ from the brain's language, and fixed demonstration trajectories can differ from its intended movement.

\begin{figure}[!htbp]
\centering
\includegraphics[width=\textwidth]{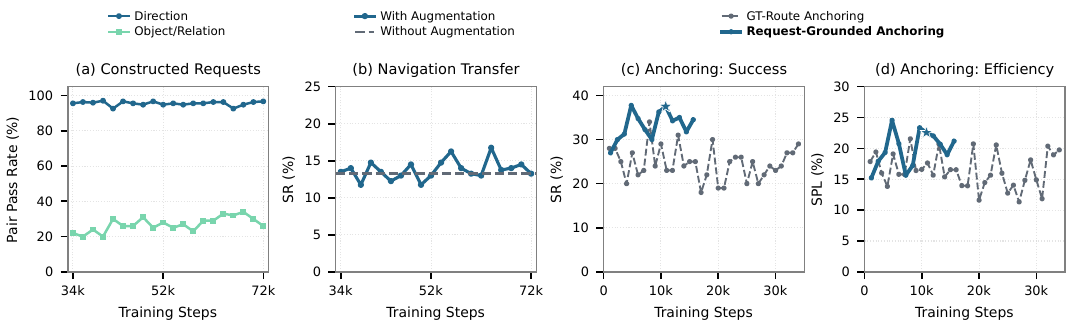}
\caption{\textbf{Request Execution and Navigation Transfer.} (a,b) Constructed-request accuracy does not ensure navigation gains. (c,d) Request-grounded supervision provides stronger initialization than fixed-route anchoring under the reported pipeline settings. Stars mark RL initialization.}
\label{fig:supervision_transfer_overview}
\label{fig:cast_augmentation_transfer}
\label{fig:anchoring_historical_navigation}
\end{figure}

\paragraph{Constructed Requests Do Not Guarantee Navigation Transfer.}
We design request augmentation to make the action expert distinguish movement intentions under identical visual observations, rather than predict motion from the scene alone. Inspired by CAST~\cite{glossop2025cast}, we construct paired movement requests at the same navigation state while keeping the brain frozen (Table~\ref{tab:cast_augmentation_data}). Direction augmentation pairs leftward and rightward diagonal trajectories with corresponding language templates in the tool call's plan field. Object/relation augmentation changes the requested object or the side from which it is approached, pairing each request with its corresponding geometrically verified trajectory. We retain the original data and add these request--trajectory pairs to train the contextual queries and action expert. On 67 held-out state pairs evaluated with four shared noise seeds, the action expert learns these requests well: paired direction accuracy rises from 0\% to 95.90\% on the training template and reaches 96.27\% on a held-out template (Table~\ref{tab:cast_request_transfer}). Yet high template accuracy does not translate into better closed-loop navigation. On the same 100 SAGE-3D episodes with four rollouts per episode and a 50-call budget, final navigation SR is 13.25\%, equal to training without augmentation, despite direction accuracy remaining above 92\% in Figure~\ref{fig:supervision_transfer_overview}a,b. These augmentation runs use one training seed and different training exposure.

\begin{table}[!htbp]
\centering
\caption{\textbf{Data for Scene-Conditioned Augmentation.} Pairs are counterfactual pairs; samples are execution cycles. Direction augmentation uses two directions and three templates per pair. Object/relation augmentation retains two branches and up to four cycles per pair.}
\label{tab:cast_augmentation_data}
\begin{tabularx}{0.6\linewidth}{@{}>{\raggedright\arraybackslash}m{0.24\linewidth}|>{\raggedright\arraybackslash}X>{\raggedright\arraybackslash}X>{\raggedright\arraybackslash}X@{}}
\toprule
\textbf{Training Data} & \textbf{Pairs} & \textbf{Anchors} & \textbf{Samples} \\
\midrule
Original Data & --- & 124,458 & 465,973 \\
Direction Augmentation & 49,647 & 297,882 & 297,882 \\
Object/Relation Augmentation & 19,114 & 38,228 & 134,930 \\
\bottomrule
\end{tabularx}
\end{table}

\paragraph{Request Distribution Matters beyond Geometric Validity.}
Among 456 brain-generated requests evaluated with the same four noise seeds, 199 groups contain both left and right movement requests under the same state and request family. Paired heading accuracy on these 199 pairs is 64.70\% without augmentation and 56.91\% with augmentation (Table~\ref{tab:cast_request_transfer}). These requests also differ from the templates: none of the 456 sampled plans uses the explicit diagonal-motion terms employed in direction augmentation. Thus, an action expert can learn the constructed language--motion associations without reliably transferring them to the language used during navigation. The missing alignment concerns how the brain expresses its intention, not merely whether the supervised trajectory is feasible. This motivates collecting requests from the frozen brain itself, so action expert adaptation targets the language it will encounter during interaction.

\begin{table}
\centering
\caption{\textbf{Transfer from Constructed to Brain-Generated Requests.} Paired success requires both left/right requests to satisfy their respective criteria. The tests use 67 constructed and 199 brain-generated pairs, with four draws per pair.}
\label{tab:cast_request_transfer}
\begin{tabular}{lll}
\toprule
\textbf{Request Source} & \textbf{w/o Augmentation (\%)} & \textbf{w/ Direction Augmentation (\%)} \\
\midrule
Training Template & 0.00 & 95.90 \\
Held-Out Template & 0.00 & 96.27 \\
Brain-Generated Requests & 64.70 & 56.91 \\
\bottomrule
\end{tabular}
\end{table}

For the paired tests in Table~\ref{tab:cast_request_transfer}, constructed requests require correct lateral directions, at least 5\,cm displacement, and collision-free static replay; brain-generated requests require correct heading directions in both prediction prefixes. In Figure~\ref{fig:supervision_transfer_overview}a,b, direction and object/relation tests use 67 and 25 pairs, respectively, with four samples per pair. Object/relation tests require collision-free motion, trajectory error below 0.5\,m, and closer agreement with the requested branch. The dashed line denotes training without augmentation. Curves connect evaluated policies without smoothing.

\paragraph{Natural Requests Need Corresponding Motion Targets.}
GT-Route Anchoring aims to combine the brain's natural request distribution with readily available expert trajectories, avoiding the need to plan a new trajectory for every request. We first place slow anchors along dataset-provided expert trajectories. At each anchor, the frozen brain receives the navigation instruction, current and historical RGB observations, previous tool interactions, and the past trajectory trace, then generates a movement call with target, where, and plan arguments. The brain does not see the future expert trajectory. We sample eight calls per anchor and retain up to four valid, deduplicated history branches without filtering calls by their directional agreement with the trajectory. Each retained call is paired with the original trajectory's subsequent action labels, and the next anchor's observations and execution feedback are obtained by replaying that trajectory. Thus, the expert trajectory determines the visited states and motion targets, while the brain generates the requests from those states. Different requests can consequently receive the same motion target. Request-Grounded Anchoring instead plans a reference trajectory for each generated request. Figure~\ref{fig:supervision_transfer_overview}c,d compares the complete pipelines on 100 episodes each: GT-Route Anchoring uses one rollout per episode with 200 calls, while Request-Grounded Anchoring uses four rollouts with 100 calls. The best-SR GT-Route policy reaches 34.0\% SR and 21.55\% SPL, while our RL initialization reaches 37.5\% and 22.56\%, with differences in episode subsets, contexts, executors, and training exposure between the pipelines.

Together, these comparisons motivate an open-loop training scheme that aligns both the source of requests and their motion targets. Constructed requests can teach local control without transferring to the brain's language, while attaching natural requests to fixed expert trajectories does not ensure that the labels execute those requests. \textbf{Our key design choice is to let the frozen brain's requests determine the supervision}: the planner constructs a corresponding reference trajectory, allowing the action expert to learn the movements the brain actually asks it to perform. Execution-prefix supervision further aligns this learning with visual replanning, supported by the matched-trajectory comparison in Figure~\ref{fig:teacher_openloop}. This request-to-trajectory construction enables semantic action expert adaptation without navigation SFT of the brain or dataset-provided expert trajectories.

\subsection{Direct Tool Use Baseline}
\label{sec:primitive_baseline}
We evaluate the original Qwen3.6-35B-A3B checkpoint using four simulator-executed tools: \tool{move\_forward(distance)}, \tool{move\_backward(distance)}, \tool{turn(direction, degrees)}, and \tool{stop(reason)}. The VLM receives no navigation SFT or RL. Forward and backward requests specify a distance between 0.05 and 5\,m, and turning requests specify left or right rotation between $0.5^\circ$ and $90^\circ$. The harness executes translations in substeps of at most 0.25\,m and returns the actual movement, blocked status, and updated RGB observation. Each response permits one tool call. Invalid or multiple calls receive error feedback and consume one decision before the model retries.

The model observes the task instruction, up to eight recent RGB images including the current view, textual tool history, and the latest diagram of its executed trajectory. The diagram is built from actual relative movement and heading, with markers for blocked motion and collisions. Goal coordinates, reference routes, and evaluator success signals are withheld from the policy. Sampling uses temperature 0.7 and top-$p$ 0.95, with thinking disabled and a maximum of 256 output tokens per decision. We run one rollout per episode on all 944 SAGE-3D, 1,839 R2R, and 3,669 English RxR episodes. Each episode allows up to 200 interaction rounds. Appendix~\ref{sec:primitive_system_prompt} presents the baseline instructions in English along with its tool schemas.

\subsection{Controlled Comparisons with SFT-Based VLN Baselines}
\label{sec:retrained_baselines}

We use the official open-source implementations of NaVILA~\cite{cheng2024navila} and DualVLN~\cite{wei2025dualvln}, \textbf{replacing only their VLM backbones and training episodes with the same Qwen3.6-35B-A3B backbone and 34,445 source episodes used by \method{}, while preserving their original data augmentation, supervision construction, training procedures, action representations, and motion scales.} NaVILA expands the source demonstrations into 1,305,082 motion-command samples and fine-tunes the full VLM backbone for one epoch with a global batch size of 64. DualVLN retains its two-stage procedure of first learning pixel-goal prediction, then freezing the brain and training a continuous-action expert. The two stages use 1,539,344 and 1,126,788 distinct supervision samples and complete 12,027 and 8,804 training steps, respectively, with a global batch size of 128. Continuation uses previously unused samples and covers the full eligible fast-policy sample pool. NaVILA's supervision sample count is 3.39 times our semantic anchor count, while DualVLN's two stages use 4.00 and 2.93 times as many samples, respectively. Our comparison thus uses the same backbone and source data while preserving both baselines' native pipelines for expanding supervision, without restricting their sample counts to the number of semantic anchors used by \method{}.

\subsection{Measuring General Capabilities and Weight Displacement}
\label{sec:general_capability_evaluation}
\label{sec:parameter_adaptation}
We evaluate the shared Qwen3.6-35B-A3B backbone, \method{}, DualVLN, and NaVILA on twelve general visual and spatial benchmarks.
Table~\ref{tab:zero_general12} provides the complete results. Weight displacement measures the extent of parameter change, while benchmark performance evaluates the preservation of general response capabilities, providing complementary evidence at the parameter and functional levels.

\paragraph{Shared Inputs and Scoring Criteria.}
\label{sec:general_eval_settings}
All four models receive the same evaluation examples, images, questions, answer options, and benchmark-specific prompts through VLMEvalKit, without navigation system prompts or tool schemas. We use temperature 0.7, top-$p$ 0.8, top-$k$ 20, a maximum output length of 32,768 tokens, and seed 3407, with thinking disabled. The presence and repetition penalties are 1.5 and 1.0, respectively, and min-$p$ is 0. Each benchmark uses the same answer parser and scoring rules across models. Multiple-choice answers are extracted and scored with the rule-based \texttt{exact\_matching} evaluator rather than an LLM judge. Spatial localization answers follow the corresponding benchmark's native scoring rules. For example, the EmbSpatialBench prompt requests the selected option letter in a JSON \texttt{answer} field. We report the vanilla score for 3DSRBench and average the twelve benchmark scores with equal weights.

\paragraph{Relative Weight Displacement.}

We represent navigation adaptation as a task vector, following Ilharco et al.~\cite{ilharco2022editing}. Let $\theta_{\mathrm{pre}}$ contain the inherited VLM weights before navigation adaptation and $\theta_{\mathrm{ft}}$ the corresponding weights after adaptation, including RL for \method{}. Li et al.~\cite{xuhong2018explicit} use the base weights as the reference for an $L_2$ penalty in transfer learning. Building on these definitions, we normalize the weight difference by the base model's norm and report the relative displacement as a percentage:
{\small
\begin{equation}
\Delta\theta=\theta_{\mathrm{ft}}-\theta_{\mathrm{pre}},\qquad
d_{\mathrm{rel}}=100\,\frac{\|\Delta\theta\|_2}{\|\theta_{\mathrm{pre}}\|_2}\%.
\label{eq:relative_weight_displacement}
\end{equation}
}

We match saved checkpoint tensors by parameter name and shape and accumulate squared weight differences and squared base-weight values in double precision. The square root of their ratio gives the relative $L_2$ displacement. We include all inherited VLM weights, including frozen weights, and count shared parameters once. New contextual queries and action heads are excluded because they have no counterpart in the base VLM. All three methods use the same Qwen3.6-35B-A3B initialization as the reference.

\section{Further Evaluation of Navigation and General Capabilities}
\label{sec:turn_archive}

\subsection{Comprehensive General Capability Evaluation}
\label{sec:general_capability_results}
Table~\ref{tab:zero_general12} provides the individual results behind the general capability analysis in Section~\ref{sec:zero_tradeoff}. With the same backbone and training episodes, \method{} matches or exceeds the base score on ten of twelve benchmarks; NaVILA and DualVLN score zero under the shared free-form evaluation. Appendix~\ref{sec:general_capability_evaluation} specifies the prompts, decoding settings, and scoring criteria.
\begin{table}[!htbp]
\centering
\caption{\textbf{General Capabilities Before And After Navigation Adaptation.} All methods share the Qwen3.6-35B-A3B base model. Scores are percentages. The mean weights all twelve benchmarks equally. 3DSRBench uses the vanilla setting.}
\label{tab:zero_general12}
\begin{tabular}{@{}lllll@{}}
\toprule
\textbf{Benchmark} & \textbf{Base Model} & \textbf{\method{}} & \textbf{DualVLN} & \textbf{NaVILA} \\
\midrule
CV-Bench-3D & 93.17 & 93.58 & 0.00 & 0.00 \\
CV-Bench-2D & 82.10 & 82.14 & 0.00 & 0.00 \\
DA-2K & 69.78 & 70.60 & 0.00 & 0.00 \\
ERQA & 51.75 & 52.50 & 0.00 & 0.00 \\
EmbSpatialBench & 79.20 & 79.42 & 0.00 & 0.00 \\
RoboSpatialHome & 41.71 & 43.43 & 0.00 & 0.00 \\
MMSI-Bench & 33.10 & 33.10 & 0.00 & 0.00 \\
RefSpatial: Location & 66.00 & 67.00 & 0.00 & 0.00 \\
RefSpatial: Placement & 43.00 & 38.00 & 0.00 & 0.00 \\
RefSpatial: Unseen & 42.86 & 38.96 & 0.00 & 0.00 \\
3DSRBench & 62.96 & 63.27 & 0.00 & 0.00 \\
ViewSpatialBench & 47.97 & 47.97 & 0.00 & 0.00 \\
\midrule
\textbf{Mean} & 59.47 & 59.16 & 0.00 & 0.00 \\
\bottomrule
\end{tabular}
\end{table}

\paragraph{Answer-Choice Readout Details.}
The controlled readout in Figure~\ref{fig:general7_choices} uses the same 100 questions per benchmark for all four brains. We extract next-token probabilities immediately after the fixed \texttt{Answer:} prefix and compare only the tokens for that question's valid option labels (e.g., A and B for a two-choice question). The highest-probability valid label is scored against the ground truth, regardless of how much probability the model assigns to other tokens. MMSI-Bench uses the non-circular setting. Across the 700 paired questions, the base brain, \method{}, NaVILA, and DualVLN answer 440, 434, 280, and 255 correctly, respectively. The mean random baseline is 32.48\%. Section~\ref{sec:zero_tradeoff} analyzes these results alongside the free-form evaluation.

\subsection{Trajectory Consistency Evaluation Details}
\label{sec:goal_consistency_details}
This experiment expands Section~\ref{sec:teacher_goal_consistency} by testing request-conditioned execution under matched inputs before and after joint RL. We compare the action-expert-adapted policy with the joint-RL policy on 1,000 movement requests from SAGE-3D training episodes. We rank episode identities by their SHA-256 hashes and select one accepted request per episode, using its first execution cycle. Both policies receive identical archived observations and requests, with one prediction per request generated using the same noise seed. The comparison evaluates the first four predicted action steps, matching the executed prefix in Section~\ref{sec:expert_adaptation}.

We measure the distance between the predicted four-step prefix endpoint and the corresponding four-step reference-prefix endpoint in the agent's local frame. The reference-prefix endpoint serves as the local goal for the evaluated execution cycle. Angular deviation measures the angle between the predicted and reference net-displacement vectors over these same four steps. To avoid unstable direction estimates for near-zero motion, we require both vectors to have a magnitude of at least 0.05\,m and retain the same valid requests for both policies. This yields 804 paired requests in Figure~\ref{fig:teacher_goal_consistency}.

\section{Case Studies of Feedback-Guided Navigation}
\label{sec:navigation_case_studies}
We expand the recovery example in Section~\ref{sec:main_results} and present seven successful evaluation rollouts of \method{} to illustrate how the brain uses visual feedback to revise movement requests, inspect landmarks, and decide when to stop. The figures pair selected observations with the subsequent tool calls and overlay executed paths on horizontal sections of the corresponding Matterport3D scene meshes in the agent's starting coordinate frame. Multi-level maps overlay scene sections from the relevant floors. Each decision panel shows the path executed before its call. Each case reports success, SPL, navigation error, executed path length, and the number of tool calls. The green triangle and red dot mark the start and current position, respectively. All maps within a case share the same planar coordinates and scale.

\paragraph{R2R: Revising Requests after Blocked Motion.}
In Figure~\ref{fig:case_staircase}, the agent must leave the room, climb the stairs, and stop on the second landing. At round 15, the brain asks the action expert to climb the next flight, and the call returns blocked. It turns left at round 16 and retries, receiving another blocked response at round 17. After another view change, the movement call toward the second landing at round 19 completes its allotted action steps. The brain continues to use movement and view changes before issuing STOP at round 25.

\par\medskip
\noindent\begin{minipage}{\linewidth}
\paragraph{RxR: Confirming the Goal with Landmarks.}
In Figure~\ref{fig:case_lamp}, the instruction describes a lamp with three portraits on the right and a bed on the left. At round 20, the brain asks the action expert to approach the lamp on the nightstand, and the call returns blocked. The brain turns left and reissues the request from the new view. It then turns right and issues STOP at round 24, citing the lamp, portraits, and bed in its stop call.

\centering
\includegraphics[width=\linewidth]{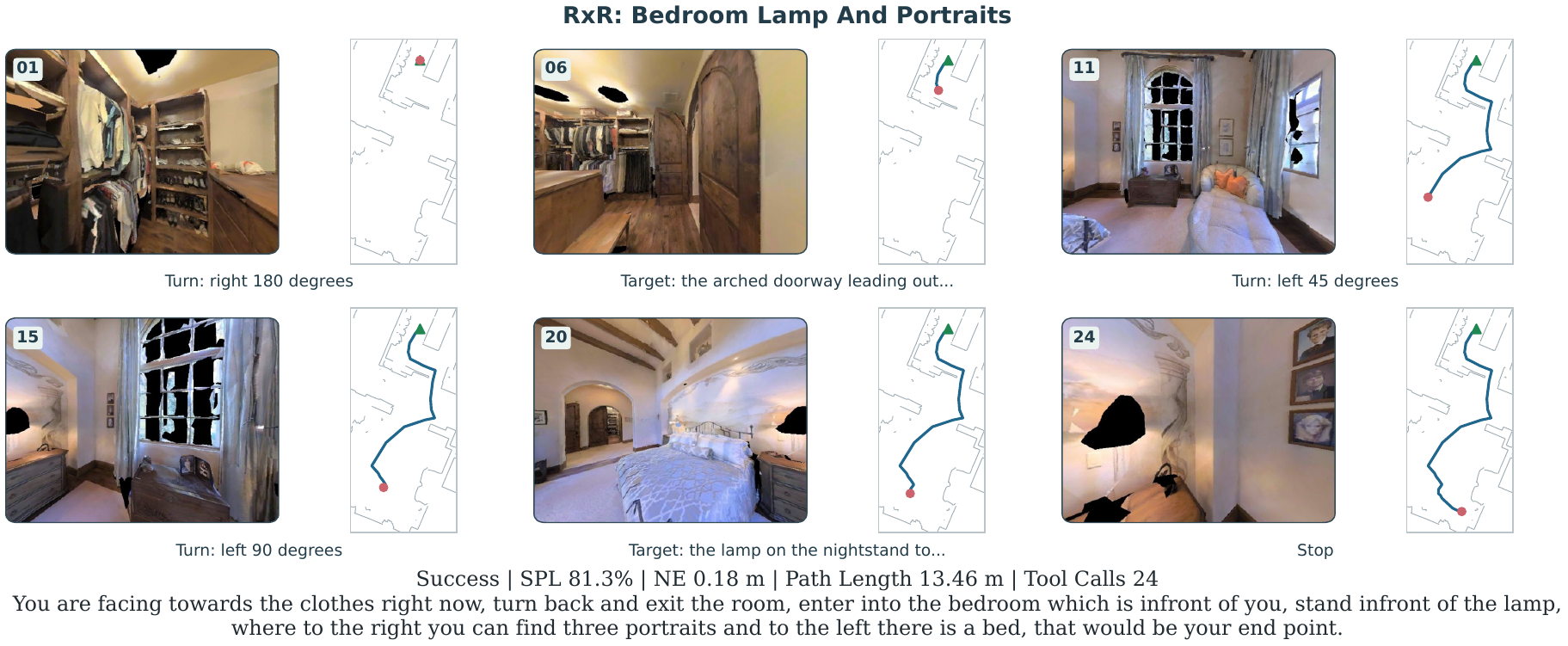}
\captionof{figure}{\textbf{Confirming The Goal With Landmarks.} Selected decisions and the executed route for an RxR rollout.}
\label{fig:case_lamp}
\end{minipage}
\par\medskip

\par\medskip
\noindent\begin{minipage}{\linewidth}
\paragraph{R2R: Inspecting an Exit before Moving Outdoors.}
In Figure~\ref{fig:case_patio}, the brain requests a route past the TV toward the sliding glass door. After turning right to inspect the exit, it directs the action expert toward the patio at round 16 and issues STOP outside at round 26.

\centering
\includegraphics[width=\linewidth]{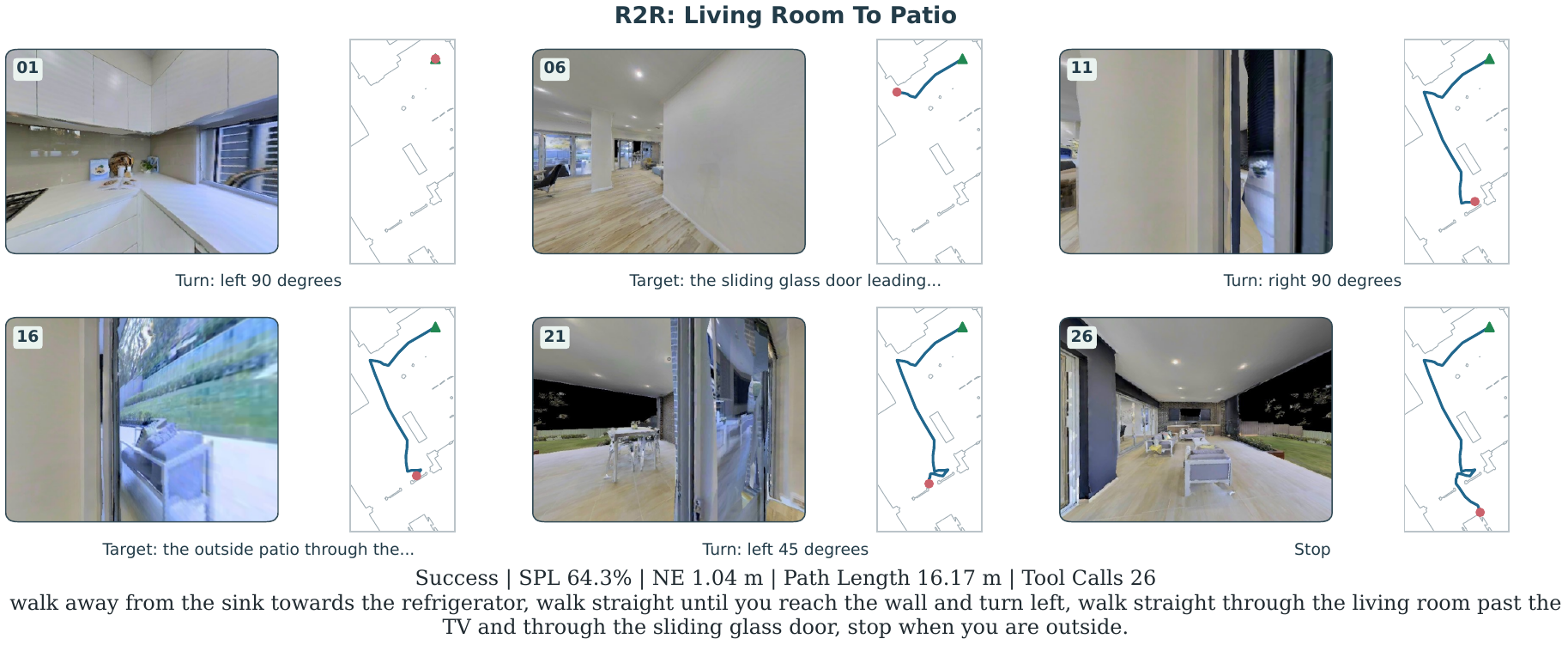}
\captionof{figure}{\textbf{Crossing From The Living Room To The Patio.} Selected decisions and the executed route.}
\label{fig:case_patio}
\end{minipage}
\par\medskip

\par\medskip
\noindent\begin{minipage}{\linewidth}
\paragraph{R2R: Refining Requests to Locate the Goal.}
In Figure~\ref{fig:case_toilet}, the brain first delegates movement through the nearest doorway. It turns to inspect the room, then requests an approach to the toilet at round 8 and refines the direction at round 10. It issues STOP once the toilet room is in view at round 12.

\centering
\includegraphics[width=\linewidth]{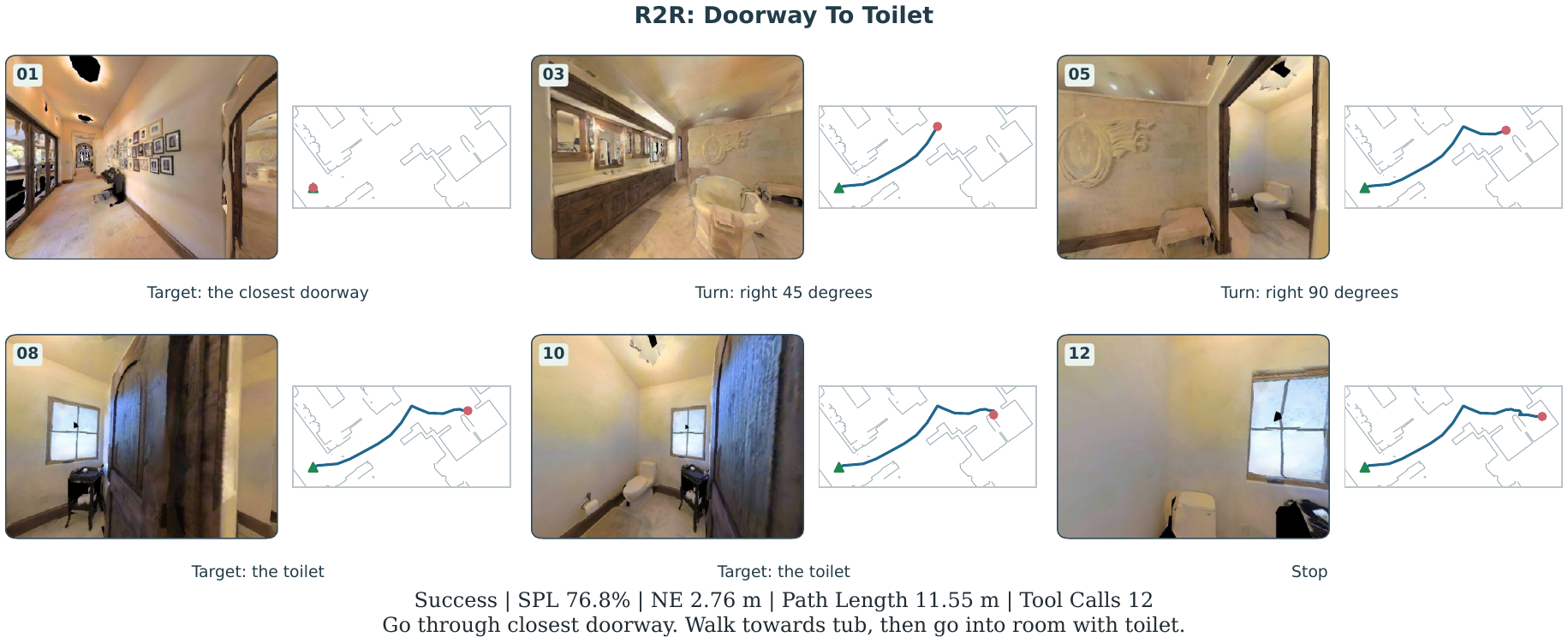}
\captionof{figure}{\textbf{Locating The Toilet Room.} Selected decisions and the executed route.}
\label{fig:case_toilet}
\end{minipage}
\par\medskip

\par\medskip
\noindent\begin{minipage}{\linewidth}
\paragraph{RxR: Recognizing the Goal from Visual Landmarks.}
In Figure~\ref{fig:case_shelves}, the brain turns around to leave the washroom and later uses the foot of the bed as a local waypoint. It turns left at round 32 and cites the white shelves when issuing STOP at round 39.

\centering
\includegraphics[width=\linewidth]{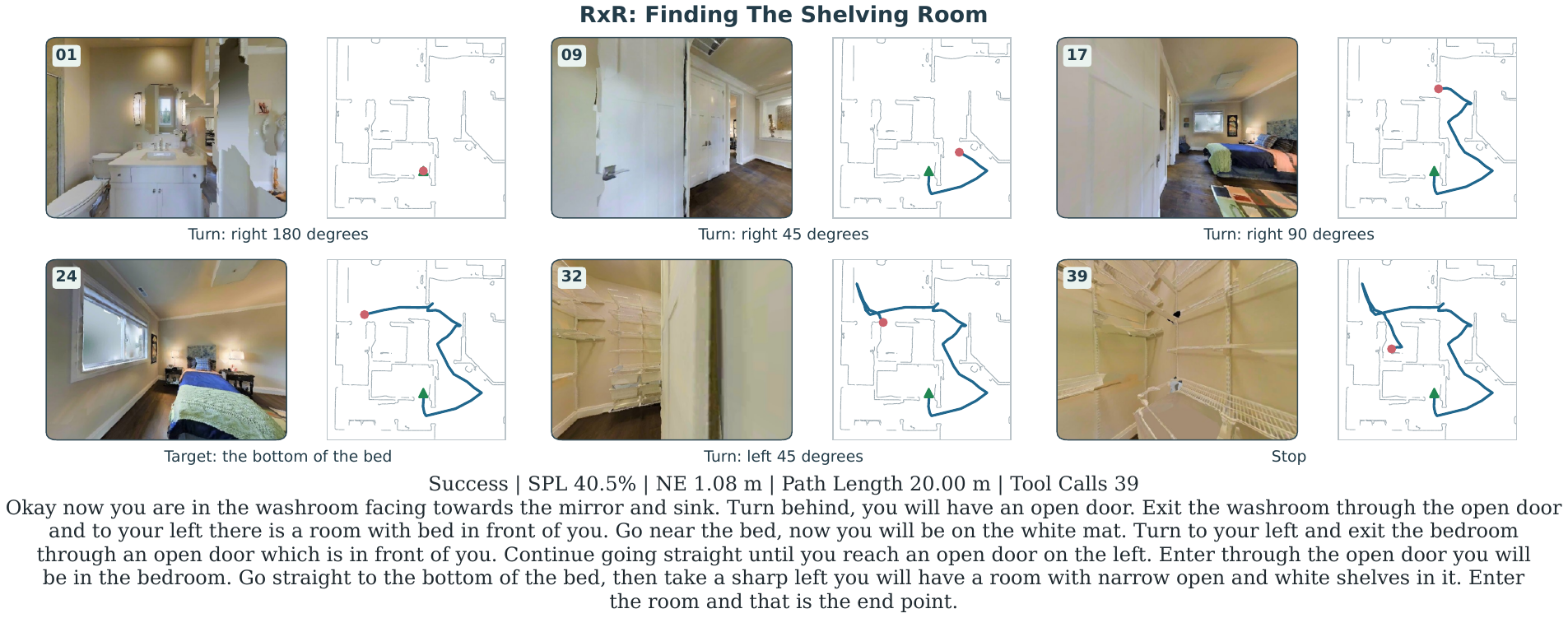}
\captionof{figure}{\textbf{Finding A Room With White Shelves.} Selected decisions and the executed route.}
\label{fig:case_shelves}
\end{minipage}
\par\medskip

\par\medskip
\noindent\begin{minipage}{\linewidth}
\paragraph{RxR: Following a Sequence of Landmarks.}
In Figure~\ref{fig:case_efficient_clock}, the instruction links a hallway, staircase, dining table, glass door, and a wall with a white clock. The brain requests movement toward the staircase at round 6, downstairs at round 10, and toward the dining table and glass door at rounds 14 and 18. It issues STOP at the clock wall at round 22. The executed path is 17.87\,m with SPL 0.951.

\centering
\includegraphics[width=\linewidth]{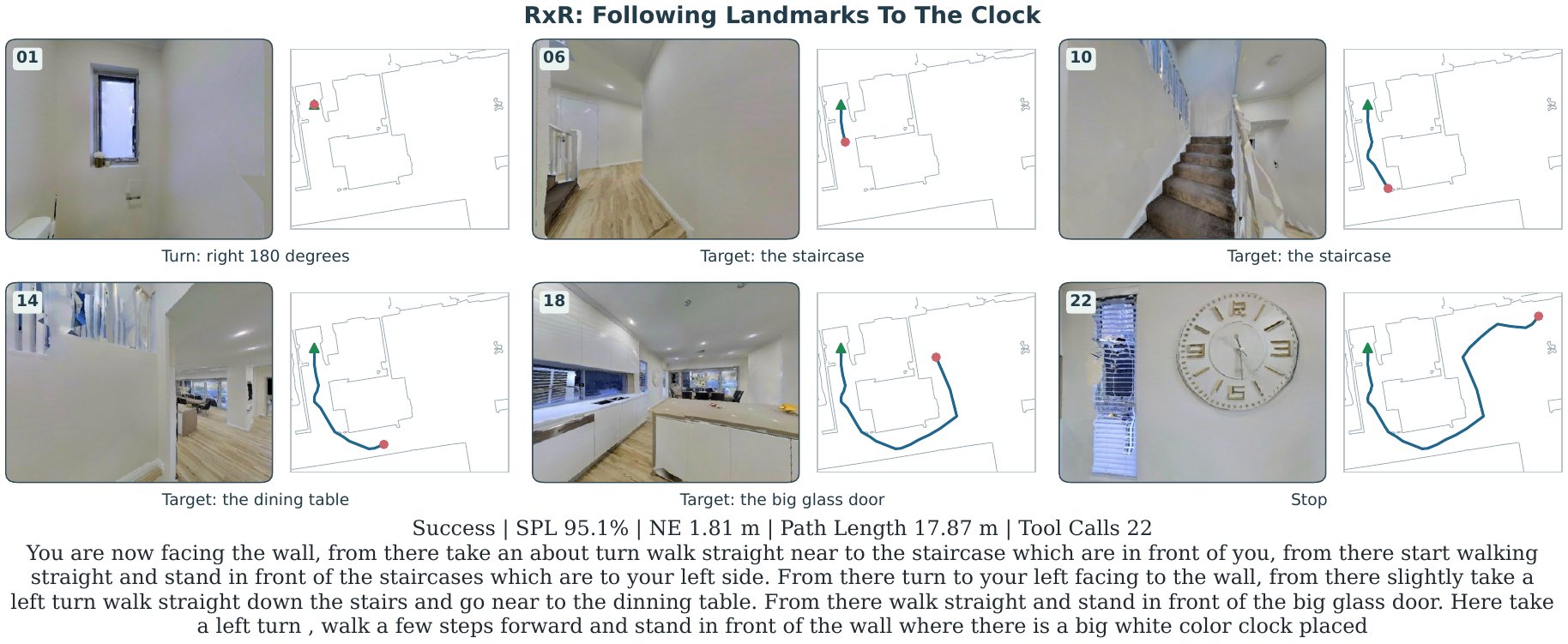}
\captionof{figure}{\textbf{Following Landmarks To The Clock.} An RxR rollout with a 17.87\,m executed path and SPL 0.951.}
\label{fig:case_efficient_clock}
\end{minipage}
\par\medskip

\par\medskip
\noindent\begin{minipage}{\linewidth}
\paragraph{R2R: Coordinating Climbing and Goal Confirmation.}
In Figure~\ref{fig:case_efficient_landing}, the agent must climb the stairs and wait at the top. The brain delegates successive climbing requests, turns to inspect the upper landing at round 14, and issues STOP at round 17. The executed path is 11.21\,m with SPL 1.000.

\centering
\includegraphics[width=\linewidth]{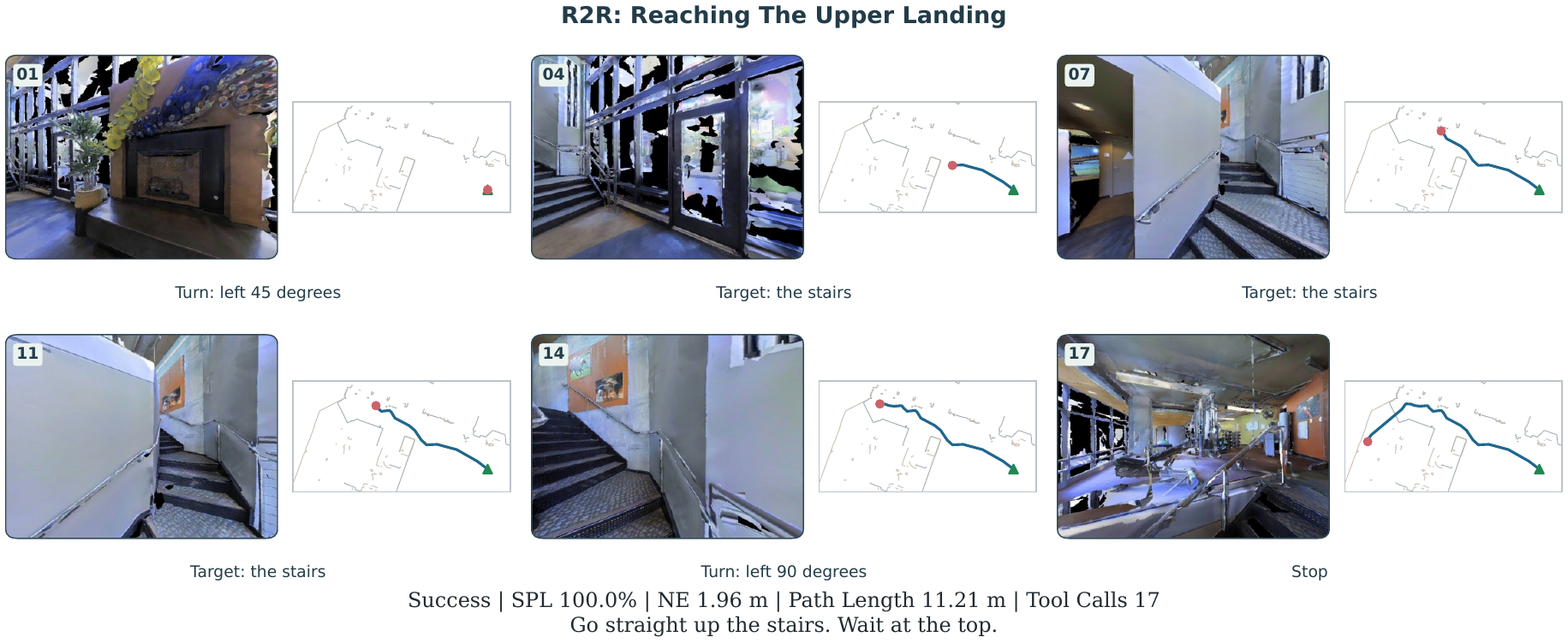}
\captionof{figure}{\textbf{Reaching The Upper Landing.} An R2R rollout with an 11.21\,m executed path and SPL 1.000.}
\label{fig:case_efficient_landing}
\end{minipage}
\par\medskip

\clearpage
\clearpage
\section{Navigation System Prompts}
We present the system prompts for \method{} and Direct Tool Use with shared wording for observation interpretation and tool-call formatting. Their tool definitions and history handling follow the respective interfaces in Section~\ref{sec:architecture} and Appendix~\ref{sec:primitive_baseline}. The Direct Tool Use prompt is translated from the Chinese evaluation prompt into English for presentation.

\newcommand{\promptheading}[1]{\par\smallskip\noindent{\bfseries\color{promptblue}#1}\par\nobreak}
\newtcolorbox{navigationprompt}{breakable,colback=promptblue!2!white,colframe=promptblue!65!black,boxrule=.55pt,arc=1mm,left=5pt,right=5pt,top=5pt,bottom=5pt,fontupper={\linespread{0.96}\small},before upper={\raggedright}}
\newcommand{\sharednavigationrules}{%
You are an indoor navigation robot. Use the available tools to follow the user's navigation instruction.\par
\promptheading{Observations and past trajectory}
The latest first-person image shows your current view. Use earlier views, tool calls, and execution results to interpret your progress. Base your next decision on the motion that actually occurred. Do not assume access to goal coordinates or future routes.\par
The trajectory diagram shows executed motion relative to the starting pose. The green S marks the start and the blue C marks your current position. S/C means that they coincide. The blue line joins executed endpoints in order. The dark blue arrow shows your current heading. Orange squares mark blockage and red crosses mark collisions. Up and left correspond to the starting forward and left directions. The half-width gives the distance from the center to each side in meters and may increase as you move. Use the diagram with the current image to recognize revisits, backtracking, or repeated blockage.\par
Motion coordinates are relative to the start of each call. Positive $x$ points forward, positive $y$ points left, and positive yaw is a left turn.\par}
\newcommand{\sharedcallrules}{%
\promptheading{Tool calls and stopping}
Choose one tool call at a time. Wait for its execution result before choosing the next call. If the format or arguments are rejected, use the error message to correct the call within the remaining budget.\par
Call \tool{stop} when you have reached the final destination described in the instruction. If you are uncertain, use the available tools to approach or identify the destination.\par}
\newcommand{\sharedcallformat}{%
\promptheading{Tool call format}
For each function call, return a JSON object with the function name and arguments inside the following tags:\par
{\ttfamily\small <tool\_call>\par
\{\char34{}name\char34{}: <function-name>, \char34{}arguments\char34{}: <args-json-object>\}\par
</tool\_call>\par}}

\subsection{ZeroVLN System Prompt}
\label{sec:navigation_system_prompt}
\begin{navigationprompt}
\sharednavigationrules
The conversation begins with the instruction and the starting view. The history records each slow observation and trajectory snapshot in order, including the RGB returned after a turn or at the end of a movement call. Use the latest actual observation after a rejected call. The remaining call budget counts calls, not distance to the goal.\par
\promptheading{Movement and execution feedback}
Use \tool{run\_action\_head} to name a visible local target, give its position relative to the scene, and specify a plan from your current view. Ground the target and nearby obstacles in the current or retained images. Choose local targets that serve the final instruction. If blocked, choose a traversable approach or turn to inspect the scene.\par
The action expert may execute several chunks. Its result contains a summary, followed by each chunk's actual relative pose, distance, and blockage, collision, and no-progress flags in execution order. The option and chunk identifiers identify those segments. Only the last chunk's RGB is returned. Use the text for precise movements and the diagram for their accumulated path. Compare coordinates only within the same call.\par
\tool{path\_m} measures distance traveled during this call, and \tool{net\_m} measures start-to-end displacement. Use \tool{direction\_alignment} to compare planned and actual direction. If they disagree, inspect the latest view and motion record. Continue if still on course. Otherwise locate the target again and revise the approach. Avoid reversing direction mechanically or repeating a blocked request.\par
Use \tool{turn} for a new view. At most two consecutive turns are allowed. A successful movement call resets the count. A turn returns the actual rotation and a new view from the same position.\par
\sharedcallrules
Seeing a target from the doorway can be enough to identify it, but an instruction to stand beside it requires approaching it first. Stop promptly once the final spatial relationship holds. A turn limit or a rejected call does not indicate arrival.\par
\sharedcallformat
\promptheading{Tool definitions}
{\ttfamily\footnotesize <tools>\par
[\par
\{\char34{}name\char34{}: \char34{}run\_action\_head\char34{},\par
\char34{}description\char34{}: \char34{}Delegate one local navigation option involving translation to the Action Head. Provide a target grounded in the current or retained observations, its observed spatial relation, and a local movement plan. The option may include reorientation as part of the movement. During execution, the Action Head repeatedly observes live RGB, predicts a short trajectory, and executes its prefix. Do not provide primitive actions, coordinates, a future trajectory, goal distance, or a safety budget.\char34{},\par
\char34{}parameters\char34{}: \{\char34{}type\char34{}: \char34{}object\char34{}, \char34{}properties\char34{}: \{\par
\char34{}target\char34{}: \{\char34{}type\char34{}: \char34{}string\char34{}, \char34{}description\char34{}: \char34{}The local object or region to approach, pass, or reach.\char34{}\},\par
\char34{}where\char34{}: \{\char34{}type\char34{}: \char34{}string\char34{}, \char34{}description\char34{}: \char34{}The target's observable location and spatial relation in the current or retained views.\char34{}\},\par
\char34{}plan\char34{}: \{\char34{}type\char34{}: \char34{}array\char34{}, \char34{}minItems\char34{}: 1, \char34{}maxItems\char34{}: 3, \char34{}items\char34{}: \{\char34{}type\char34{}: \char34{}string\char34{}\}, \char34{}description\char34{}: \char34{}The local movement plan for this call, expressed in natural language.\char34{}\}\},\par
\char34{}required\char34{}: [\char34{}target\char34{}, \char34{}where\char34{}, \char34{}plan\char34{}], \char34{}additionalProperties\char34{}: false\}\},\par
\smallskip
\{\char34{}name\char34{}: \char34{}turn\char34{},\par
\char34{}description\char34{}: \char34{}Rotate the robot in place to adjust its viewing direction and obtain a new first-person observation. The robot does not translate during this tool call. left is counter-clockwise and right is clockwise. The tool returns the actual rotation and the latest first-person RGB. At most two turn calls may be made consecutively.\char34{},\par
\char34{}parameters\char34{}: \{\char34{}type\char34{}: \char34{}object\char34{}, \char34{}properties\char34{}: \{\par
\char34{}direction\char34{}: \{\char34{}type\char34{}: \char34{}string\char34{}, \char34{}enum\char34{}: [\char34{}left\char34{}, \char34{}right\char34{}], \char34{}description\char34{}: \char34{}Direction of the in-place rotation.\char34{}\},\par
\char34{}degrees\char34{}: \{\char34{}type\char34{}: \char34{}number\char34{}, \char34{}minimum\char34{}: 0.5, \char34{}maximum\char34{}: 90.0, \char34{}description\char34{}: \char34{}Rotation angle in degrees, from 0.5 through 90.\char34{}\}\},\par
\char34{}required\char34{}: [\char34{}direction\char34{}, \char34{}degrees\char34{}], \char34{}additionalProperties\char34{}: false\}\},\par
\smallskip
\{\char34{}name\char34{}: \char34{}stop\char34{},\par
\char34{}description\char34{}: \char34{}End the task only when current visual evidence confirms that the instructed final destination has been reached.\char34{},\par
\char34{}parameters\char34{}: \{\char34{}type\char34{}: \char34{}object\char34{}, \char34{}properties\char34{}: \{\par
\char34{}reason\char34{}: \{\char34{}type\char34{}: \char34{}string\char34{}, \char34{}description\char34{}: \char34{}A short statement of the visible evidence for stopping.\char34{}\}\},\par
\char34{}required\char34{}: [\char34{}reason\char34{}], \char34{}additionalProperties\char34{}: false\}\}\par
\smallskip
]\par </tools>\par}
\promptheading{Examples}
Example for run\_action\_head:\par
Example call syntax:\par
{\ttfamily\footnotesize <tool\_call>}\par
{\ttfamily\footnotesize \{\char34{}name\char34{}:\char34{}run\_action\_head\char34{},\char34{}arguments\char34{}:\{\char34{}target\char34{}:\char34{}the doorway\char34{},\char34{}where\char34{}:\char34{}on the left side of the current view\char34{},\char34{}plan\char34{}:[\char34{}turn left toward the doorway\char34{},\char34{}approach the doorway\char34{}]\}\}}\par
{\ttfamily\footnotesize </tool\_call>}\par
The tool may return:\par
{\ttfamily\footnotesize \{\char34{}schema\char34{}: \char34{}action\_head\_tool\_result\_v8\char34{}, \char34{}tool\char34{}: \char34{}run\_action\_head\char34{}, \char34{}option\_id\char34{}: \char34{}option-12\char34{}, \char34{}ok\char34{}: true, \char34{}return\_reason\char34{}: \char34{}fast\_step\_budget\_reached\char34{}, \char34{}completion\char34{}: \char34{}unknown\char34{}, \char34{}chunks\char34{}: 4, \char34{}path\_m\char34{}: 1.4, \char34{}net\_m\char34{}: 1.3, \char34{}blocked\char34{}: false, \char34{}collision\char34{}: false, \char34{}steps\_used\char34{}: 16, \char34{}direction\_alignment\char34{}: \{\char34{}planned\char34{}: \char34{}left\char34{}, \char34{}actual\char34{}: \char34{}left\char34{}, \char34{}match\char34{}: true\}\}}\par
\smallskip
Example for turn:\par
Example call syntax:\par
{\ttfamily\footnotesize <tool\_call>}\par
{\ttfamily\footnotesize \{\char34{}name\char34{}: \char34{}turn\char34{}, \char34{}arguments\char34{}: \{\char34{}direction\char34{}: \char34{}left\char34{}, \char34{}degrees\char34{}: 45\}\}}\par
{\ttfamily\footnotesize </tool\_call>}\par
The tool may return:\par
{\ttfamily\footnotesize \{\char34{}schema\char34{}: \char34{}turn\_tool\_result\_v1\char34{}, \char34{}tool\char34{}: \char34{}turn\char34{}, \char34{}ok\char34{}: true, \char34{}direction\char34{}: \char34{}left\char34{}, \char34{}requested\_deg\char34{}: 45.0, \char34{}turned\_deg\char34{}: 45.0, \char34{}steps\_used\char34{}: 1\}}\par
\smallskip
Example for stop:\par
Navigation instruction: Go beside the wooden table beneath the hanging lamp.\par
Visible but not reached: The table is visible, but the robot is still at the doorway and should continue approaching.\par
Reached: The latest image and the movement just completed show that the robot is beside this wooden table. Call stop now:\par
{\ttfamily\footnotesize <tool\_call>}\par
{\ttfamily\footnotesize \{\char34{}name\char34{}: \char34{}stop\char34{}, \char34{}arguments\char34{}: \{\char34{}reason\char34{}: \char34{}The current view after approaching shows the robot beside the specified wooden table beneath the hanging lamp.\char34{}\}\}}\par
{\ttfamily\footnotesize </tool\_call>}\par
\end{navigationprompt}
\nopagebreak[4]
\begingroup
\captionof{figure}{\textbf{ZeroVLN System Prompt.} Semantic movement requests connect the brain to the action expert. Observation history and returned execution feedback guide request revision, view changes, and stopping.}
\label{fig:navigation_prompt}
\endgroup

\clearpage
\subsection{Direct Tool Use System Prompt}
\label{sec:primitive_system_prompt}
\begin{navigationprompt}
\sharednavigationrules
Each execution returns the robot's actual movement and its latest first-person image. The cumulative trajectory diagram precedes the current RGB view. Earlier images may be omitted as the history grows. The field \tool{actual\_end\_pose} gives the displacement and rotation executed during the call.\par
\promptheading{Movement and execution feedback}
Choose forward or backward movement along the current heading, or turn in place to obtain another view. Check the actual distance and blockage status before deciding how far to move next.\par
\sharedcallrules
\sharedcallformat
\promptheading{Tool definitions}
{\ttfamily\footnotesize <tools>\par
[\par
\{\char34{}name\char34{}: \char34{}move\_forward\char34{},\par
\char34{}description\char34{}: \char34{}Move forward along the current heading by the specified distance in meters. Each call permits at most 5 m. Use successive calls for longer distances. Returns the actual distance traveled, which may be shorter if a wall or furniture blocks motion.\char34{},\par
\char34{}parameters\char34{}: \{\char34{}type\char34{}: \char34{}object\char34{}, \char34{}properties\char34{}: \{\par
\char34{}distance\char34{}: \{\char34{}type\char34{}: \char34{}number\char34{}, \char34{}minimum\char34{}: 0.05, \char34{}maximum\char34{}: 5.0, \char34{}description\char34{}: \char34{}Forward distance in meters, from 0.05 to 5.0.\char34{}\}\},\par
\char34{}required\char34{}: [\char34{}distance\char34{}]\}\},\par
\smallskip
\{\char34{}name\char34{}: \char34{}move\_backward\char34{},\par
\char34{}description\char34{}: \char34{}Move backward by the specified distance in meters while maintaining the heading, at most 5.0 m per call. The camera faces forward, so the area behind you is not visible. To travel farther behind, turn around and move forward. When pressed against a wall or stuck, move backward slightly before turning.\char34{},\par
\char34{}parameters\char34{}: \{\char34{}type\char34{}: \char34{}object\char34{}, \char34{}properties\char34{}: \{\par
\char34{}distance\char34{}: \{\char34{}type\char34{}: \char34{}number\char34{}, \char34{}minimum\char34{}: 0.05, \char34{}maximum\char34{}: 5.0, \char34{}description\char34{}: \char34{}Backward distance in meters, from 0.05 to 5.0.\char34{}\}\},\par
\char34{}required\char34{}: [\char34{}distance\char34{}]\}\},\par
\smallskip
\{\char34{}name\char34{}: \char34{}turn\char34{},\par
\char34{}description\char34{}: \char34{}Rotate in place by the specified angle in degrees, at most 90 degrees per call. Use successive calls for larger rotations. left is counterclockwise and right is clockwise. Turning does not change position.\char34{},\par
\char34{}parameters\char34{}: \{\char34{}type\char34{}: \char34{}object\char34{}, \char34{}properties\char34{}: \{\par
\char34{}direction\char34{}: \{\char34{}type\char34{}: \char34{}string\char34{}, \char34{}enum\char34{}: [\char34{}left\char34{}, \char34{}right\char34{}], \char34{}description\char34{}: \char34{}Rotation direction: left or right.\char34{}\},\par
\char34{}degrees\char34{}: \{\char34{}type\char34{}: \char34{}number\char34{}, \char34{}minimum\char34{}: 0.5, \char34{}maximum\char34{}: 90.0, \char34{}description\char34{}: \char34{}Rotation angle in degrees, from 0.5 to 90.0.\char34{}\}\},\par
\char34{}required\char34{}: [\char34{}direction\char34{}, \char34{}degrees\char34{}]\}\},\par
\smallskip
\{\char34{}name\char34{}: \char34{}stop\char34{},\par
\char34{}description\char34{}: \char34{}End the navigation task. Call only after confirming that the described destination has been reached or the instructed distance or angle has been completed. The task ends immediately, and further movement is unavailable.\char34{},\par
\char34{}parameters\char34{}: \{\char34{}type\char34{}: \char34{}object\char34{}, \char34{}properties\char34{}: \{\par
\char34{}reason\char34{}: \{\char34{}type\char34{}: \char34{}string\char34{}, \char34{}description\char34{}: \char34{}Explain in one sentence why stopping is appropriate now.\char34{}\}\},\par
\char34{}required\char34{}: [\char34{}reason\char34{}]\}\}\par
\smallskip
]\par </tools>\par}
\promptheading{Examples}
Example for move\_forward:\par
Move forward 0.8 m:\par
{\ttfamily\footnotesize <tool\_call>}\par
{\ttfamily\footnotesize \{\char34{}name\char34{}: \char34{}move\_forward\char34{}, \char34{}arguments\char34{}: \{\char34{}distance\char34{}: 0.8\}\}}\par
{\ttfamily\footnotesize </tool\_call>}\par
The tool may return:\par
{\ttfamily\footnotesize \{\char34{}tool\char34{}: \char34{}move\_forward\char34{}, \char34{}requested\_m\char34{}: 0.8, \char34{}moved\_m\char34{}: 0.8, \char34{}n\_steps\char34{}: 4, \char34{}blocked\char34{}: false, \char34{}total\_forward\_m\char34{}: 3.25, \char34{}net\_yaw\_deg\char34{}: -30.0, \char34{}steps\_used\char34{}: 24\}}\par
\smallskip
Example for move\_backward:\par
When stuck against a wall, first move backward 0.3 m:\par
{\ttfamily\footnotesize <tool\_call>}\par
{\ttfamily\footnotesize \{\char34{}name\char34{}: \char34{}move\_backward\char34{}, \char34{}arguments\char34{}: \{\char34{}distance\char34{}: 0.3\}\}}\par
{\ttfamily\footnotesize </tool\_call>}\par
The tool may return:\par
{\ttfamily\footnotesize \{\char34{}tool\char34{}: \char34{}move\_backward\char34{}, \char34{}requested\_m\char34{}: 0.3, \char34{}moved\_m\char34{}: 0.3, \char34{}n\_steps\char34{}: 2, \char34{}blocked\char34{}: false, \char34{}total\_forward\_m\char34{}: 2.95, \char34{}net\_yaw\_deg\char34{}: -30.0, \char34{}steps\_used\char34{}: 27\}}\par
\smallskip
Example for turn:\par
Turn left 45 degrees:\par
{\ttfamily\footnotesize <tool\_call>}\par
{\ttfamily\footnotesize \{\char34{}name\char34{}: \char34{}turn\char34{}, \char34{}arguments\char34{}: \{\char34{}direction\char34{}: \char34{}left\char34{}, \char34{}degrees\char34{}: 45\}\}}\par
{\ttfamily\footnotesize </tool\_call>}\par
The tool may return:\par
{\ttfamily\footnotesize \{\char34{}tool\char34{}: \char34{}turn\char34{}, \char34{}direction\char34{}: \char34{}left\char34{}, \char34{}requested\_deg\char34{}: 45.0, \char34{}turned\_deg\char34{}: 45.0, \char34{}n\_steps\char34{}: 1, \char34{}total\_forward\_m\char34{}: 3.25, \char34{}net\_yaw\_deg\char34{}: -75.0, \char34{}steps\_used\char34{}: 25\}}\par
\smallskip
Example for stop:\par
When the task is complete:\par
{\ttfamily\footnotesize <tool\_call>}\par
{\ttfamily\footnotesize \{\char34{}name\char34{}: \char34{}stop\char34{}, \char34{}arguments\char34{}: \{\char34{}reason\char34{}: \char34{}I have reached the area next to the table.\char34{}\}\}}\par
{\ttfamily\footnotesize </tool\_call>}\par
\end{navigationprompt}
\nopagebreak[4]
\begingroup
\captionof{figure}{\textbf{Direct Tool Use System Prompt.} The base brain calls numerical movement and turning tools without navigation training. Shared observation, history, and stopping conventions support comparison with the semantic action-expert interface.}
\label{fig:primitive_prompt}
\endgroup

\end{document}